\documentclass[10pt,twocolumn,letterpaper]{article}

\usepackage[pagenumbers]{iccv} % To force page numbers, e.g. for an arXiv version
\usepackage{multirow}

\definecolor{iccvblue}{rgb}{0.21,0.49,0.74}
\usepackage[pagebackref,breaklinks,colorlinks,allcolors=iccvblue]{hyperref}

\def\paperID{14428} % *** Enter the Paper ID here
\def\confName{ICCV}
\def\confYear{2025}

\title{To Label or Not to Label: PALM – A Predictive Model \\ for Evaluating Sample Efficiency in Active Learning Models}

\author{Julia Machnio \\
Pioneer Centre for AI \\
University of Copenhagen \\
{\tt\small juma@di.ku.dk}
\and
Mads Nielsen \\
Pioneer Centre for AI \\
University of Copenhagen \\
{\tt\small madsn@di.ku.dk}
\and
Mostafa Mehdipour Ghazi \\
Pioneer Centre for AI \\
University of Copenhagen \\
{\tt\small ghazi@di.ku.dk}
}

\begin{document}
\maketitle

\begin{abstract}

Active learning (AL) seeks to reduce annotation costs by selecting the most informative samples for labeling, making it particularly valuable in resource-constrained settings. However, traditional evaluation methods, which focus solely on final accuracy, fail to capture the full dynamics of the learning process. To address this gap, we propose PALM (Performance Analysis of Active Learning Models), a unified and interpretable mathematical model that characterizes AL trajectories through four key parameters: achievable accuracy ($\text{A}_{\max}$), coverage efficiency ($\delta$), early-stage performance ($\alpha$), and scalability ($\beta$). PALM provides a predictive description of AL behavior from partial observations, enabling the estimation of future performance and facilitating principled comparisons across different strategies. We validate PALM through extensive experiments on CIFAR-10/100 and ImageNet-50/100/200, covering a wide range of AL methods and self-supervised embeddings. Our results demonstrate that PALM generalizes effectively across datasets, budgets, and strategies, accurately predicting full learning curves from limited labeled data. Importantly, PALM reveals crucial insights into learning efficiency, data space coverage, and the scalability of AL methods. By enabling the selection of cost-effective strategies and predicting performance under tight budget constraints, PALM lays the basis for more systematic, reproducible, and data-efficient evaluation of AL in both research and real-world applications. The code is available at: \url{https://github.com/juliamachnio/PALM}.

\end{abstract}

\section{Introduction}
\label{sec:intro}

\begin{figure}
    \centering
    \includegraphics[width=0.88\linewidth]{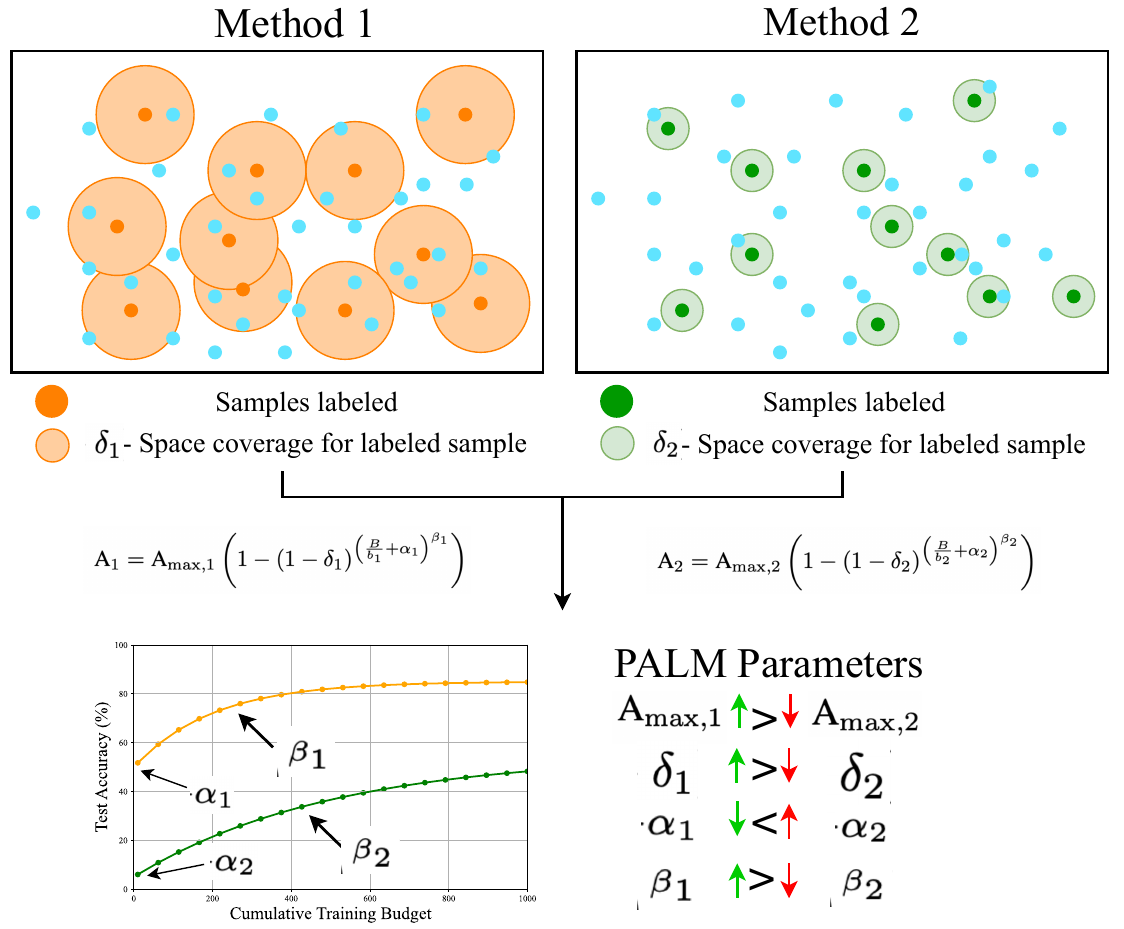}
    \caption{Illustration of the PALM method for comparing AL strategies through interpretable parameters. Methods 1 and 2 differ in how labeled samples (green and orange) cover the data space, with $\delta$ representing the average coverage per labeled sample. The corresponding PALM equations describe accuracy growth as a function of the cumulative annotation budget ($B$), with parameters controlling early-stage performance ($\alpha$), scalability ($\beta$), and maximum achievable accuracy ($\text{A}_{\max}$). The learning curves show how different parameter configurations affect AL dynamics: Method 1 achieves faster accuracy gains and higher asymptotic performance due to better coverage efficiency and scalability.}
    \label{fig:enter-label}
\end{figure}

With the rapid growth of large-scale, domain-specific datasets in real-world applications, the cost and effort required for data labeling have become significant bottlenecks in machine learning pipelines. In many domains, obtaining labeled data necessitates substantial human expertise, time, and resources, making it impractical to fully-annotate large datasets. Active learning (AL) addresses this challenge by iteratively selecting the most informative samples for labeling, enabling models to achieve high performance with fewer labeled examples \cite{settles2009active}. This makes AL particularly attractive in settings where labeled data is scarce or expensive, such as medical imaging \cite{wang2024comprehensive,wu2023active,chen2024think}, autonomous driving \cite{greer2024and,lin2024exploring}, and anomaly detection in IoT systems \cite{zakariah2023anomaly}.

Although AL provides a solution to reduce annotation costs, its effectiveness in deep learning largely depends on the sample selection strategy. Over the years, a variety of AL methods have been proposed \cite{xu2003representative,voevodski2012active,aghaee2016active,zhang2023labelbench}, typically categorized into three main groups based on their selection criteria: uncertainty-based methods, diversity-based methods, and core-set approaches. Uncertainty-based methods prioritize samples where the model exhibits low confidence \cite{lewis1995sequential,scheffer2001active,gissin2019discriminative,nguyen2022measure}. Diversity-based methods aim to maximize data space coverage by selecting diverse samples \cite{hu2010off,ash2019deep,kee2018query,jin2022one}. Core-set approaches focus on selecting representative subsets that best capture the global data distribution \cite{sener2017active,liang2024semantic,maalouf2022unified}. Recently, a fourth category, typicality-based methods, has emerged, particularly effective in scenarios with limited annotation budgets (the number of labeled samples). These methods select samples that are most representative of the underlying data distribution, aiming to enhance coverage and learning efficiency \cite{hacohen2022active,yehuda2022active,mishal2024dcom}.

Integrating self-supervised learning (SSL) into active learning pipelines has further enhanced typicality-based methods by providing richer feature representations that improve both sample selection and model initialization \cite{yehuda2022active,mishal2024dcom,hacohen2022active}. SSL techniques such as SimCLR \cite{chen2020simple}, BYOL \cite{grill2020bootstrap}, or MoCov2+ and MoCov3 \cite{chen2020improved, chen2021empirical} have become powerful tools for learning transferable representations from unlabeled data. The synergy between AL and SSL has proven especially beneficial under limited annotation budgets, where high-quality feature representations help identify the most informative and representative samples.

Despite these advancements, the performance of AL methods remains highly sensitive to experimental settings, including model initialization, dataset properties, and task complexity \cite{beluch2018power}. Furthermore, current evaluation protocols typically focus on the final accuracy achieved after a fixed number of iterations, neglecting critical factors such as early learning efficiency, spatial coverage, and the scalability of learning gains over time. Recent work has highlighted the need for fair comparisons of AL methods in various scenarios but continues to rely on metrics such as accuracy and area under the ROC curve (AUC), which provide only a limited view of overall performance \cite{mittal2025realistic}.

These comparison limitations reveal a fundamental gap; there is no general predictive model for evaluating and comparing AL performance across diverse methods, datasets, and annotation budgets. In real-world scenarios, where annotation resources are finite, the objective is not only to achieve high accuracy but also to ensure robustness and generalization within limited budgets \cite{settles2009active}. Moreover, deep learning models often retain predictive ability in uncovered regions of the data space, meaning that performance depends not only on labeled coverage but also on the model's inherent generalization capacity. Thus, evaluating active learning strategies requires considering both coverage expansion and the model's ability to learn from unlabeled regions. Without standardized, predictive evaluations incorporating these factors, selecting the most effective AL strategy remains a costly and inconsistent process.

To address the challenge of fair comparison, we propose \textit{PALM}, a mathematical model for the Performance Analysis of Active Learning Models, describing the dynamics of AL processes through a compact set of interpretable parameters as shown in Figure \ref{fig:enter-label}. PALM captures not only how accuracy grows with increasing labeled data but also how uncovered regions contribute to overall performance through generalization. It estimates the full learning curve from a limited number of iterations and predicts outcomes such as final accuracy, data coverage, early learning efficiency, and scalability. This formulation generalizes across different AL methods, datasets, and configurations and is compatible with or without SSL-based feature representations. By providing an interpretable and predictive description of AL behavior, PALM enables practitioners to estimate the number of labels required to achieve target performance and identify the most cost-effective strategies under resource constraints.

Our main contributions are as follows: (1) we propose \textit{PALM}, a simple yet powerful mathematical model that describes the dynamics of AL and predicts key performance metrics from limited labeled data; (2) we empirically evaluate PALM’s ability to compare different AL methods, datasets, and configurations without requiring full dataset annotation; (3) we demonstrate that PALM can estimate final accuracy, data space coverage, early learning efficiency, and scalability of learning gains across various AL settings; (4) we analyze the impact of incorporating SSL embeddings into AL pipelines, showing that PALM remains effective with or without SSL-based representations; and (5) we provide a practical approach for estimating the number of labeling samples needed to achieve target performance, facilitating real-world deployments under strict budget constraints.

\section{Methods}
\label{sec:covering}

In this section, we present the underlying calculations of PALM and the properties that enable it to predict final accuracy and space coverage for active learning methods. Our formulation is grounded in fundamental probabilistic principles \cite{ross2014introduction}, considering the random distribution of objects in a space and their contribution to coverage estimation. This approach is inspired by the random covering problem \cite{janson1986random}, as well as concepts such as covering and effective sample size \cite{cui2019class} and dynamic coverage and margin mix \cite{mishal2024dcom}.

\subsection{Expected Coverage Fraction}

\textbf{Definition 1} (Coverage Probability).
Let $\mathbb{X}\subset\mathbb{R}^d$ represent a $d$-dimensional space, and let $\{O_i\}_{i=1}^s$ denote a collection of $s$ randomly placed objects within $\mathbb{X}$. Each object $O_i$ is a measurable subset of $\mathbb{X}$. For a given point $x \in \mathbb{X}$, we define the coverage probability $P_C(x)$ as the probability that $x$ is covered by at least one object:
\begin{equation}
    P_C(x) = 1 - P_{UC}(x),
\end{equation}
where $P_{UC}(x)$ represents the probability that $x$ is not covered by any object in $\{O_i\}_{i=1}^s$.

\textbf{Definition 2} (Coverage Probability with $s$ Independent Objects).
Assume that each object $O_i$ is placed independently and covers a given point $x \in \mathbb{X}$ with the same probability $p$. The probability that a single object does not cover $x$ is $1 - p$. Since the objects are independent, the probability that no object covers $x$ is given by:
\begin{equation}
    P_{UC}(x) = (1 - p)^s.
\end{equation}
Therefore, the overall coverage probability of point $x$ is:
\begin{equation}
\label{def2}
    P_C(x) = 1 - (1 - p)^s.
\end{equation}
As $s \to \infty$, $P_C(x)$ approaches 1, ensuring that $x$ is almost surely covered. When $s = 0$, we have $P_C(x) = 0$. Moreover, since $(1 - p)^s$ decreases with increasing $s$ for $0 < p < 1$, the probability of non-coverage diminishes as the number of objects increases.

\textbf{Corollary 1} (Boundary Behavior of the Coverage Probability).\label{Corollary 1}
The coverage probability $P_C(x)$, as defined in \cref{def2}, exhibits different boundary behaviors. As the number of objects grows infinitely, the probability that $x$ is covered approaches 1, ensuring full coverage of the space:
\begin{equation}
    \lim_{s \to \infty} P_C(x) = 1 - \lim_{s \to \infty} (1 - p)^s,
\end{equation}
and since $0 < 1 - p < 1$, we have:
\begin{equation}
    \lim_{s \to \infty} (1 - p)^s = 0.
\end{equation}
Thus, as the number of objects approaches infinity, the coverage probability becomes:
\begin{equation}
    \lim_{s \to \infty} P_C(x) = 1.
\end{equation}
Conversely, as the number of objects approaches zero, the probability of coverage also approaches zero, indicating no coverage of the space:
\begin{equation}
    \lim_{s \to 0^+} P_C(x) = 1 - \lim_{s \to 0^+} (1 - p)^s = 0.
\end{equation}

\textbf{Definition 3} (Expected Coverage Fraction).
Let $\mathbb{X}$ be a space with finite volume $V$, and let $\delta = V_{O_i} / V$ denote the fraction of the total volume covered by a single randomly placed object $O_i$, assuming all objects have the same volume and are placed independently and uniformly over $\mathbb{X}$. Under the assumption of independence, the expected fraction of $\mathbb{X}$ covered by $s$ objects, denoted by $\mathbb{E}_\text{C}$, is given by:
\begin{equation} 
    \mathbb{E}_\text{C} = 1 - (1 - \delta)^s
\end{equation}
where $0 \leq \delta \leq 1$. Note that $\mathbb{E}_\text{C}$ describes the expected proportion of the entire space covered and is independent of any specific point in $\mathbb{X}$. This equation exhibits the same limiting behavior as outlined in the previous corollary. A larger $\delta$ implies that fewer objects are needed to achieve significant coverage of $\mathbb{X}$ since each object covers a greater proportion of the space.

\subsection{Active Learning Pools}

To define the properties of AL methods, we adopt the notation introduced in \cite{mishal2024dcom}, with a modification in the interpretation of $\delta$. Specifically, in this context, $\delta$ represents the fraction of the space $\mathbb{X}$ covered by a single labeled sample rather than the radius of a covering ball.

% Let $P$ denote the underlying probability distribution over the data space $\mathbb{X}$.
% Assume the existence of a true labeling function $f: \mathbb{X} \to \mathbb{Y}$.
We define $\mathbb{L} \subseteq \mathbb{X}$ as the labeled set of points and $\mathbb{U} \subseteq \mathbb{X}$ as the unlabeled set, such that $\mathbb{X} = \mathbb{L} \cup \mathbb{U}$, with the cardinality of the labeled set $|\mathbb{L}| = B$, representing the cumulative annotation budget collected iteratively from the AL pool. A labeling round refers to an iteration of the active learning process, in which a subset of points from $\mathbb{U}$ is selected for labeling, and the labeled set $\mathbb{L}$ is updated accordingly. Throughout the AL process, the cumulative budget $B \leq |\mathbb{X}|$ increases as more rounds are completed.
% Besides, we define $b$ as the mean number of labeling samples per round. 

\textbf{Definition 4} (Expected Coverage in Active Learning Pools).  
The expected fraction of $\mathbb{X}$ covered by $B$ labeled samples, selected independently, is given by:
\begin{equation} 
\label{def4}
    \mathbb{E}_\text{C} = 1 - (1 - \delta)^B,
\end{equation}
where, $\delta$ captures the average contribution of a single labeled sample to the overall coverage of $\mathbb{X}$, which inherently depends on the AL strategy used to select the samples. Specifically, more effective strategies prioritize samples that maximize coverage, leading to higher $\delta$ values. Hence, comparing $\delta$ across different strategies allows for assessing their relative effectiveness in covering the space.

Additionally, the accuracy of a model trained on the labeled set $\mathbb{L} \subseteq \mathbb{X}$ depends not only on the fraction of the space covered by labeled samples but also on the quality of the predictions within both the covered and uncovered regions \cite{yehuda2022active}. Furthermore, the uncovered regions contribute to the model's generalization ability by influencing its behavior in areas with sparse or no labeled data.

\textbf{Definition 5} (Accuracy as a Function of Coverage Probability).
The overall test accuracy of a trained model on the labeled set can be expressed as:
\begin{equation} 
    \text{A} = \text{A}_{\text{C}} \, P_{\text{C}} + \text{A}_{\text{UC}} \, (1 - P_{\text{C}}),
\end{equation}
where $\text{A}_{\text{C}}$ is the model's accuracy in the covered regions of $\mathbb{X}$, $\text{A}_{\text{UC}}$ is its accuracy in the uncovered regions, and $P_{\text{C}}$ is the probability that a random test point lies within a covered region.  
By substituting $P_{\text{C}}$ from \cref{def4}, we obtain:
\begin{equation} 
\label{def5}
    \text{A} = \text{A}_{\text{C}} \, \left(1 - (1 - \delta)^B \right) + \text{A}_{\text{UC}} \, (1 - \delta)^B,
\end{equation}
which establishes a simple yet powerful relationship between test accuracy and the proportion of covered space. When the labeled budget $B$ is small, the overall accuracy is dominated by $\text{A}_{\text{UC}}$, as most of the space remains uncovered. As $B$ increases, coverage expands and accuracy improves, eventually converging to $\text{A}_{\text{C}}$ as $B \to \infty$. For small $\delta$ values, the accuracy exhibits exponential growth behavior, approximately following $1 - e^{-B\delta}$, capturing the rate at which coverage and accuracy improve with increasing budget.

This formulation unifies the concepts of coverage and accuracy growth, showing how increasing labeled samples enhances accuracy while recognizing the importance of uncovered regions. In real-world scenarios, where models often retain some generalization capacity in uncovered areas, $\text{A}_{\text{UC}}$ becomes a crucial factor in performance evaluation. However, it does not fully describe how uncovered regions influence the rate of accuracy improvement during AL. To address this, we introduce additional parameters to better reflect the impact of uncovered areas on learning dynamics.

\textbf{Definition 6} (Generalized Accuracy as a Function of Coverage with Exponential Adjustment).  
We define the generalized test accuracy function as:
\begin{equation} 
\label{Def6}
\text{A} = \text{A}_{\max} \left(1 - (1 - \delta)^{(B + \alpha)^\beta}\right),
\end{equation}
where $\text{A}_{\max}$ is the maximum achievable accuracy, $\alpha > -B$ is a shifting parameter that adjusts the effective starting point of learning by capturing the contribution of uncovered regions, and $\beta > 0$ is a scaling parameter that controls the rate at which accuracy increases with the number of labeled samples. These parameters introduce greater flexibility in modeling how space coverage shapes the learning outcome. An analysis of the asymptotic behavior of this function is provided in Supplementary Materials, showing how $\alpha$ and $\beta$ influence initial accuracy, growth rate, and the impact of uncovered regions on learning dynamics.

The generalized accuracy function offers a robust model for approximating AL training curves across varying initial conditions, AL strategies, and datasets. Given accuracy observations at four or more distinct cumulative budgets $B$, the parameters \( \text{A}_{\max} \), \( \delta \), \( \alpha \), and \( \beta \) can be estimated by solving a system of equations using nonlinear regression methods. The computational complexity of this estimation is approximately $\mathcal{O}(\log(B))$ when employing optimized exponentiation techniques.

More straightforward optimization approaches typically exhibit higher complexity, approaching $\mathcal{O}(B^\beta)$ for large $B$. Moreover, the rapid growth of the term \( (B + \alpha)^\beta \) can lead to large numerical values, increasing the risk of floating-point precision errors. To mitigate these issues and enhance numerical stability, it is beneficial to normalize the cumulative budget $B$ by the mean annotation budget per iteration, denoted $b$. This normalization reduces computational complexity to approximately $\mathcal{O}((B/b)^\beta)$ and aligns the model with the iterative structure of AL, where labeled data is acquired in fixed increments of size $b$ per iteration.

\textbf{Definition 7} (Normalized Accuracy Function for Active Learning Budget Scaling).  
The generalized accuracy function with normalized budget $B$ is defined as:
\begin{equation}
\label{def7}
    \text{A} = \text{A}_{\max} \left(1 - (1 - \delta)^{\left(\frac{B}{b} + \alpha\right)^\beta} \right),
\end{equation}
where $b$ is the mean number of labeled samples per iteration. This normalization aligns the accuracy function with the iterative structure of AL and mitigates numerical instability associated with large budget values. Estimating the characteristic parameters (\( \delta, b, \alpha, \beta, \text{A}_{\max} \)) from observed learning curves enables direct and meaningful comparisons between different AL methods.

\textbf{Theorem 1} (Comparison of Active Learning Methods Using the Normalized Accuracy Function).  
Consider two active learning methods 1 and 2 described by their respective accuracy functions:
\begin{align}
\text{A}_1 &= \text{A}_{\max, 1} \left(1 - (1 - \delta_1)^{\left(\frac{B}{b_1} + \alpha_1\right)^{\beta_1}} \right), \\
\text{A}_2 &= \text{A}_{\max, 2} \left(1 - (1 - \delta_2)^{\left(\frac{B}{b_2} + \alpha_2\right)^{\beta_2}} \right).
\end{align}
For a given budget $B$, Method 1 outperforms Method 2 if $\text{A}_1 > \text{A}_2$.  
Importantly, $\text{A}_{\max}$ determines the asymptotic performance as $B \to \infty$, with the method having higher $\text{A}_{\max}$ achieving superior long-term accuracy. The rate of accuracy improvement, influenced by $\delta$, $b$, $\alpha$, and $\beta$, further captures the efficiency of each method in reaching its asymptotic potential. The full proof is presented in Supplementary Materials.

\textbf{Lemma 1} (Comparison Criterion for Active Learning Methods).  
For any budget $B$, two active learning methods, A and B, can be compared directly through:
\begin{equation}
\text{A}_1 > \text{A}_2.
\end{equation}
This provides a simple and effective criterion to identify the better-performing method for a given annotation budget, making it easier to evaluate performance across different stages of the AL process.

\textbf{Corollary 2} (Comparison of Learning Efficiency and Parameter Influence).
Assuming both methods achieve the same maximum accuracy, their performance can be distinguished through the following considerations:
\begin{itemize}
\item \textbf{Coverage Efficiency} ($\delta$): A higher $\delta$ indicates that each labeled sample provides more effective coverage of the space, enhancing overall efficiency.
\item \textbf{Budget Efficiency} ($b$): A smaller $b$ results in more frequent, smaller batches of labeled data, leading to faster and more incremental accuracy improvements.
\item \textbf{Initial Learning Efficiency} ($\alpha$): A lower $\alpha$ boosts the starting accuracy, which is especially advantageous in low-budget scenarios.
\item \textbf{Scaling of Learning Gains} ($\beta$): A larger $\beta$ increases the accuracy improvement rate as more samples are labeled.
\end{itemize}
Hence, active learning methods can be quantitatively compared by evaluating these parameters and $\text{A}_{\max}$ that defines the maximum achievable accuracy as the budget increases.

Overall, this work presents a unified equation for modeling AL performance through a normalized accuracy function that captures the effects of coverage efficiency, budget scaling, and learning dynamics. By introducing coverage efficiency ($\delta$), budget efficiency ($b$), initial accuracy shift ($\alpha$), and learning gain scaling ($\beta$), the model captures the early-stage and long-term behavior of AL methods. This formulation enables a direct, quantitative comparison of different AL strategies based on their estimated parameters, independent of the dataset or initial conditions.

\section{Experimental Setup}
\label{sec:experiments}

We evaluate the proposed PALM model in fully supervised settings using ResNet-18 and ResNet-50 architectures \cite{he2016deep} on the CIFAR-10 and CIFAR-100 datasets \cite{krizhevsky2009learning}, as well as on subsets of ImageNet \cite{deng2009imagenet}, including ImageNet-50, ImageNet-100, and ImageNet-200. In this context, fully supervised refers to training the models exclusively on the labeled subset $\mathbb{L}$ of the dataset. At each iteration, $\mathbb{L}$ is incrementally expanded with newly labeled samples in accordance with standard AL procedures. We adapted an existing AL framework to ensure reproducibility across various AL sampling strategies \cite{munjal2022towards,hacohen2022active}. In each experiment, we estimated the PALM parameters, which characterize the AL dynamics, based on the cumulative annotation budget $B$ at the corresponding iteration.

\subsection{Supervised Training on CIFARs}

For CIFAR-10 and CIFAR-100, we trained a ResNet-18 model for at least 100 epochs. The training process employed stochastic gradient descent (SGD) with Nesterov momentum (0.9), weight decay (0.0003), and a cosine learning rate schedule, starting with an initial learning rate of 0.025. The models were trained using a batch size of 512. Standard data augmentation techniques, including random cropping and horizontal flipping, were applied. For further implementation details, please refer to \cite{munjal2022towards,mishal2024dcom}.

\subsection{Supervised Training on ImageNet Subsets}

For ImageNet-50, ImageNet-100, and ImageNet-200, we adopted the same optimization settings and data augmentation protocol as used in the CIFAR experiments, but with a ResNet-50 backbone \cite{he2016deep}.

\subsection{Self-Supervised Feature Extraction}

To obtain semantically meaningful feature representations, we trained SimCLR on each dataset using the official implementation from \cite{chen2020simple}. For CIFAR-10 and CIFAR-100, we employed a ResNet-18 encoder with a projection head mapping to a 128-dimensional vector, training for 500 epochs. The training hyperparameters followed the settings outlined in SCAN \cite{van2020scan}. After training, we extracted representations from the penultimate layer, resulting in 512-dimensional embeddings for CIFAR-10 and CIFAR-100 and 2048-dimensional embeddings for ImageNet.

For the ImageNet subsets, we also extracted features using the official pretrained models of BYOL \cite{grill2020bootstrap} (ResNet-50 trained for 300 epochs with a batch size of 512), MoCov2+ \cite{chen2020improved} (ResNet-50 trained for 800 epochs with a batch size of 256), and MoCov3 \cite{chen2021empirical} (ResNet-50 trained for 1000 epochs with a batch size of 4096). We computed the mean and standard deviation of the features extracted from the training samples, applied normalization accordingly, and used the same values to normalize the features extracted from the validation and test samples.

\subsection{Active Learning Strategies}

To demonstrate PALM's ability to approximate active learning dynamics across diverse query strategies, we evaluated several commonly used AL methods: \textit{Random Sampling}, \textit{TypiClust} \cite{hacohen2022active}, \textit{Uncertainty Sampling}, which selects samples with the lowest maximum softmax probability, \textit{Max Entropy}, which selects samples with the highest predictive entropy, \textit{Min Margin}, which selects samples with the smallest margin between the top two softmax probabilities, and \textit{DBAL} \cite{gal2017deep}. For reproducibility and fair comparison, we followed the experimental settings outlined in \cite{munjal2022towards}.

\section{Results and Discussion}

\begin{figure*}[t]
\centering
\includegraphics[width=1.75\columnwidth]{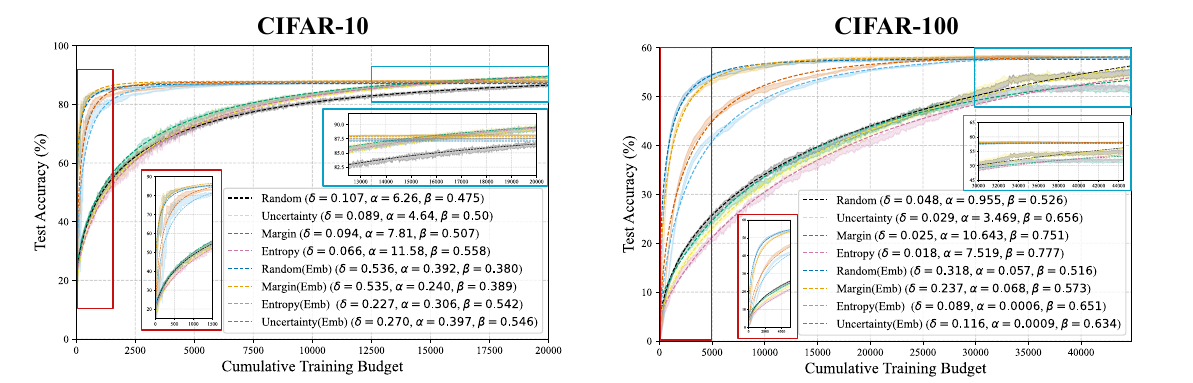}
\caption{PALM approximation of AL curves on CIFAR-10 ($b=20$) and CIFAR-100 ($b=100$). Each plot displays the accuracy trajectories of multiple AL strategies fitted with the PALM model across the entire annotation budget. Two distinct behaviors emerge in both datasets: methods with gradual accuracy growth and those utilizing pretrained embeddings that quickly reach maximum accuracy. These differences are reflected in the fitted PALM parameters. Methods using pretrained embeddings consistently show higher $\delta$ values, indicating greater sample efficiency, while non-embedding methods exhibit higher $\alpha$ values, indicating delayed learning progress. Despite the varying complexity of the datasets, PALM successfully models the learning dynamics and highlights clear differences between AL strategies. Note that the highest test accuracy for CIFAR-100 is approximately 65\%. All quantitative results are available in the Supplementary Materials.}
\label{fig:full_pool} % Optional label for referencing
\end{figure*}

\noindent\textit{\textbf{AL Behavior:}} We first evaluate the ability of PALM to approximate AL dynamics across different sampling strategies on CIFAR-10 and CIFAR-100. For CIFAR-10, we used a mean budget size of $b=20$ over 1000 iterations, while for CIFAR-100, the budget was $b=100$ over 450 iterations. Each AL strategy was executed until the unlabeled pool was nearly exhausted. For each AL curve, we estimated the PALM parameters $(\delta, \alpha, \beta, \text{A}_{\max})$ using nonlinear least squares to fit the model to the observed empirical accuracy as a function of the cumulative annotation budget $B$. To ensure stable and interpretable fits, we imposed constraints on the parameters. These constraints help reduce ambiguity in the parameter estimates. Figure \ref{fig:full_pool} presents the AL curves together with their corresponding PALM fits. Across both datasets, two distinct behaviors emerge: (1) AL strategies that achieve gradual accuracy improvements over extended annotation budgets and (2) strategies leveraging pretrained embeddings, which exhibit rapid accuracy gains and early convergence toward $\text{A}_{\max}$. These differences in behavior are reflected in the fitted PALM parameters.

The parameter $\delta$ quantifies the coverage efficiency per labeled sample. Across both CIFAR-10 and CIFAR-100, methods that employ pretrained embeddings achieve significantly higher $\delta$ values. For example, on CIFAR-10, Margin sampling without embeddings yields $\delta = 0.094$, whereas its embedding-based counterpart achieves $\delta = 0.535$. Similarly, on CIFAR-100, Random sampling improves from $\delta = 0.048$ (without embeddings) to $\delta = 0.318$ (with embeddings). These indicate that pretrained embeddings enhance the representational quality of selected samples, thereby improving space coverage and annotation efficiency.

The parameter $\alpha$ controls the initial offset of learning progress. Higher $\alpha$ values indicate a delayed onset of learning, a trend that is especially pronounced in non-embedding methods. For example, on CIFAR-100, Margin sampling without embeddings exhibits $\alpha = 10.643$, whereas its embedding-based counterpart reduces $\alpha$ to 0.068, indicating faster initial learning when using SSL-derived representations. Moreover, the parameter $\beta$ captures the scaling of accuracy gains for the annotation budget. Larger $\beta$ values correspond to steeper growth in accuracy. On CIFAR-100, Entropy sampling with embeddings demonstrates $\beta = 0.651$, the highest among the evaluated strategies, indicating rapid convergence.

\begin{figure*}[t]
\centering
\includegraphics[width=1.7\columnwidth]{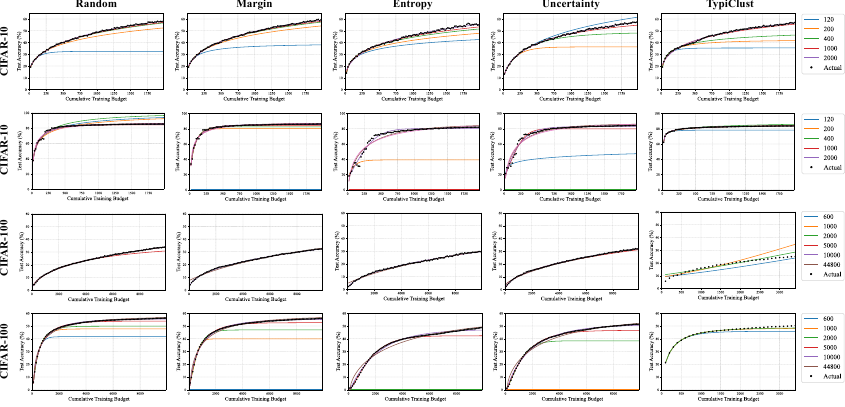}
\caption{PALM predictions of AL training curves fitted using varying numbers of cumulative budget points, as indicated in the legend. Each plot compares the predicted trajectories with the actual performance across different AL strategies on CIFAR-10 and CIFAR-100. Note that the strategies in rows 2 and 4 utilize embeddings extracted using the SimCLR method. We find that accurate predictions of the full learning dynamics can be made with cumulative budgets of approximately 1,000 samples for CIFAR-10 (2\% of the dataset) and 5,000 samples for CIFAR-100 (10\% of the dataset). In certain cases on CIFAR-10, reasonable predictions can be made with as few as 400 samples (0.8\% of the dataset), demonstrating PALM’s ability to model learning behavior even with limited early-stage supervision.}
\label{fig:estimation_cifar} % Optional label for referencing
\end{figure*}

Despite similar final accuracies across methods, the PALM parameters offer a fine-grained view of sample efficiency and learning behavior. For example, while Random sampling with embeddings achieves a high $\delta = 0.536$, Entropy sampling without embeddings is limited to $\delta = 0.066$, despite achieving comparable asymptotic performance. This illustrates that PALM not only captures endpoint accuracy but also characterizes the trajectory of learning dynamics. These findings empirically validate \cref{def7}, where $\delta$, $\alpha$, and $\beta$ jointly describe the efficiency, onset, and scaling of active learning progress. The clear distinction in behaviors across AL strategies, as revealed by the fitted parameters, confirms that PALM provides a consistent and interpretable model of AL dynamics across different datasets, methods, and embedding strategies. Full quantitative results are provided in the Supplementary Materials.

While PALM accurately fits the learning trajectories across AL strategies, we observe minor deviations in the later rounds of the CIFAR-100 experiments. This is likely due to the progressive exhaustion of informative examples in the unlabeled pool. As AL progresses, the remaining samples tend to be noisier or less representative, which increases prediction uncertainty and affects fit stability. Although PALM assumes an effectively infinite unlabeled pool, this simplification may lead to small estimation drift once the pool is nearly depleted. Nevertheless, the observed deviations remain within 2\% of test accuracy across all budget levels. Importantly, PALM is designed to support early-stage decision-making, where accurate prediction from limited partial observations is most impactful.

\noindent\textit{\textbf{Limited Budgets:}} To further assess the predictive capability of PALM, we evaluated its performance when fitted to partial learning curves, using only a limited number of annotated samples across different AL methods and datasets. The results are presented in Figure \ref{fig:estimation_cifar}. For CIFAR-10, an accurate approximation of the full learning dynamics was achieved with as few as 1,000 labeled samples, representing just 2\% of the dataset. Notably, for some strategies, high-quality predictions emerged with only 200–300 labeled samples, corresponding to approximately 0.8–1.5\% of the dataset. Methods leveraging pretrained embeddings required even fewer samples to achieve strong fits, further highlighting the efficiency gains provided by SSL-based representations. Additionally, for CIFAR-100, a more complex dataset with 100 classes, reliable approximations were obtained using 5,000–10,000 labeled samples. This translates to approximately 50–100 labeled examples per class, demonstrating that PALM can generalize effectively to higher-dimensional spaces with greater class diversity.  

TypiClust, a method specifically designed for small-budget scenarios and reliant on pretrained embeddings, demonstrated particularly strong approximation with extremely limited supervision. On CIFAR-100, PALM accurately modeled TypiClust's learning trajectory using just 1,000 labeled samples (roughly 10 examples per class), underscoring its effectiveness under resource-constrained conditions. The use of SSL-based embeddings significantly reduced the computational cost of AL iterations, offering practical advantages for large-scale or real-time annotation workflows. This finding suggests that embedding-based AL pipelines not only enhance model performance but also improve training efficiency, making them a compelling choice for the design of future AL systems.  

It is worth noting that for CIFAR-10, PALM occasionally struggles to fit the sharp knee point of the learning curve due to the dataset's relatively simple structure and the rapid accuracy gains observed in early iterations. In contrast, this effect is less pronounced on CIFAR-100, where the increased complexity and class imbalance align more closely with real-world scenarios, enabling smoother approximations across the full budget range.  

\begin{figure*}[t]
\centering
\includegraphics[width=1.7\columnwidth]{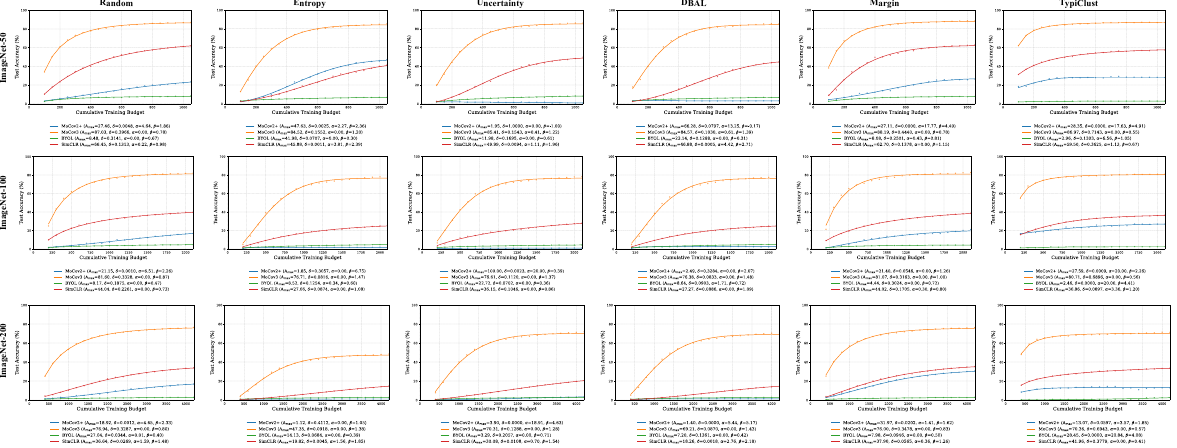}
\caption{PALM predictions of AL curves on ImageNet subsets. Each plot compares the performance of different AL strategies combined with various SSL methods (MoCov2+, MoCov3, BYOL, and SimCLR). PALM accurately models learning dynamics under extreme annotation constraints, capturing distinct behaviors across methods and embeddings. MoCov3 consistently achieves the highest achievable accuracy ($\text{A}_{\text{max}}$) and coverage efficiency ($\delta$), indicating superior representation quality and faster learning. In contrast, BYOL exhibits slow, nearly linear accuracy growth with minimal coverage, indicating delayed learning that requires larger annotation budgets. SimCLR and MoCov2+ show varying dynamics depending on the AL strategy, underscoring the sensitivity of embedding performance to sample selection methods. These findings highlight PALM's ability to model and compare AL behavior across diverse settings.}
\label{fig:img200} % Optional label for referencing
\end{figure*}

\noindent\textit{\textbf{Large-Scale Datasets:}} Furthermore, to assess the generalization capability of PALM on large-scale, natural datasets, we conducted experiments on ImageNet-50, ImageNet-100, and ImageNet-200 using a total annotation budget corresponding to 4\% of the training samples. Figure \ref{fig:img200} presents the results of this experiment. As seen, PALM successfully approximated the learning dynamics across all tested active learning strategies and SSL-based embeddings, reinforcing its applicability to realistic, resource-constrained scenarios. Beyond capturing overall learning trajectories, PALM effectively models distinct learning behaviors, including methods that achieve rapid initial improvements and those with delayed performance onset. These differences are reflected in the fitted parameters, particularly $\alpha$ and $\delta$, which control the starting point and coverage efficiency of the learning process.  

A particularly notable case arises with the BYOL embeddings. Due to BYOL’s extremely slow and nearly linear accuracy progression within the small budget, PALM fits the curve as if the method has already plateaued. This behavior occurs because the early-stage dynamics are not sufficiently captured with the limited number of labeled samples. However, signs of acceleration appear in methods like TypiClust, where the final iterations show an upward trend, suggesting that with larger annotation budgets, BYOL would eventually follow the growth patterns predicted by our mathematical model. This observation underscores the need for more labeled samples to estimate future performance for slow-starting methods reliably.  

Differences in $\beta$ across AL strategies also reveal important scalability trends, with higher $\beta$ values indicating steeper accuracy gains following the initial labeling phase. Notably, strategies combined with MoCov3 consistently exhibit higher $\beta$, reflecting their ability to capitalize on accumulated labeled data more effectively. A closer analysis of the ImageNet-200 results further illustrates these insights:  
\begin{itemize}
    \item \textbf{MoCov3} consistently achieves the highest $\text{A}_{\max}$ and $\delta$ values across all strategies, indicating superior representation quality and faster learning progression, making it highly effective under limited supervision.  
    \item \textbf{SimCLR} shows competitive $\text{A}_{\max}$ but lower $\delta$, suggesting slower per-sample gains and greater sensitivity to the choice of active learning strategy, as evidenced by the variability of $\alpha$ across methods.  
    \item \textbf{BYOL} displays consistently low $\delta$ values and, in some cases, near-zero values (e.g., with TypiClust), reflecting minimal coverage per labeled sample and delayed learning. For TypiClust specifically, a high $\alpha$ and steep $\beta$ indicate a late but accelerating learning phase, aligning with the observed rise in the final iterations.  
    \item \textbf{MoCov2+} generally lags behind the other embeddings, with lower $\text{A}_{\max}$ and $\delta$ across strategies, further highlighting the varying effectiveness of SSL embeddings in active learning pipelines.  
\end{itemize}
The findings show that PALM not only captures overall performance trends but also uncovers nuanced interactions between SSL embeddings and AL strategies. It offers a comprehensive model for comparing methods, balancing trade-offs between early performance and long-term scalability, and guiding annotation planning in practical applications.

\section{Conclusion}
\label{conclusion}

We presented PALM, a unified mathematical model for the performance analysis of active learning models by capturing the dynamics of AL curves through a compact set of interpretable parameters. Unlike conventional evaluation protocols that compare methods based solely on accuracy at fixed budgets, PALM provides a principled approach to understanding how active learning strategies behave throughout the entire annotation process. By modeling coverage efficiency ($\delta$), sampling behavior ($b$), early-stage performance ($\alpha$), scalability ($\beta$), and achievable accuracy ($\text{A}_{\max}$), PALM offers a comprehensive characterization of active learning trajectories from limited observations.

Our experiments on CIFAR and ImageNet benchmarks, covering a diverse range of AL strategies and SSL embeddings, demonstrate that PALM generalizes across datasets, methods, and budget regimes despite using only a small fraction of labeled data. It not only accurately predicts learning curves under extreme annotation constraints but also reveals key differences in how methods scale, where they accelerate, and how embedding quality interacts with sampling strategies. These insights expose the limitations of relying on final accuracy alone and establish PALM as a fair model for cost-aware comparison of AL methods.

PALM advances AL by shifting the focus from isolated performance measurements to predictive modeling of the learning process. This enables more informed, data-efficient decision-making in real-world settings, where annotation costs dominate and strategy selection is critical. By accounting for the dynamics and learning efficiency, PALM lays the basis for a new standard in AL evaluation, paving the way toward more systematic and scalable AL design.

\noindent\textbf{Acknowledgements.}
This project has received funding from the Pioneer Centre for AI, DNRF, grant number P1.

{
\small
\bibliographystyle{ieeenat_fullname}
\bibliography{main}

\begin{thebibliography}{40}
\providecommand{\natexlab}[1]{#1}
\providecommand{\url}[1]{\texttt{#1}}
\expandafter\ifx\csname urlstyle\endcsname\relax
  \providecommand{\doi}[1]{doi: #1}\else
  \providecommand{\doi}{doi: \begingroup \urlstyle{rm}\Url}\fi

\bibitem[Aghaee et~al.(2016)Aghaee, Ghadiri, and Baghshah]{aghaee2016active}
Amin Aghaee, Mehrdad Ghadiri, and Mahdieh~Soleymani Baghshah.
\newblock Active distance-based clustering using k-medoids.
\newblock In \emph{Advances in Knowledge Discovery and Data Mining: 20th
  Pacific-Asia Conference, PAKDD 2016, Auckland, New Zealand, April 19-22,
  2016, Proceedings, Part I 20}, pages 253--264. Springer, 2016.

\bibitem[Ash et~al.(2019)Ash, Zhang, Krishnamurthy, Langford, and
  Agarwal]{ash2019deep}
Jordan~T Ash, Chicheng Zhang, Akshay Krishnamurthy, John Langford, and Alekh
  Agarwal.
\newblock Deep batch active learning by diverse, uncertain gradient lower
  bounds.
\newblock \emph{arXiv preprint arXiv:1906.03671}, 2019.

\bibitem[Beluch et~al.(2018)Beluch, Genewein, N{\"u}rnberger, and
  K{\"o}hler]{beluch2018power}
William~H Beluch, Tim Genewein, Andreas N{\"u}rnberger, and Jan~M K{\"o}hler.
\newblock The power of ensembles for active learning in image classification.
\newblock In \emph{Proceedings of the IEEE conference on computer vision and
  pattern recognition}, pages 9368--9377, 2018.

\bibitem[Chen et~al.(2024)Chen, Ma, Cui, and Xia]{chen2024think}
Jiayi Chen, Benteng Ma, Hengfei Cui, and Yong Xia.
\newblock Think twice before selection: Federated evidential active learning
  for medical image analysis with domain shifts.
\newblock In \emph{Proceedings of the IEEE/CVF Conference on Computer Vision
  and Pattern Recognition}, pages 11439--11449, 2024.

\bibitem[Chen et~al.(2020{\natexlab{a}})Chen, Kornblith, Norouzi, and
  Hinton]{chen2020simple}
Ting Chen, Simon Kornblith, Mohammad Norouzi, and Geoffrey Hinton.
\newblock A simple framework for contrastive learning of visual
  representations.
\newblock In \emph{International conference on machine learning}, pages
  1597--1607. PmLR, 2020{\natexlab{a}}.

\bibitem[Chen et~al.(2020{\natexlab{b}})Chen, Fan, Girshick, and
  He]{chen2020improved}
Xinlei Chen, Haoqi Fan, Ross Girshick, and Kaiming He.
\newblock Improved baselines with momentum contrastive learning.
\newblock \emph{arXiv preprint arXiv:2003.04297}, 2020{\natexlab{b}}.

\bibitem[Chen et~al.(2021)Chen, Xie, and He]{chen2021empirical}
Xinlei Chen, Saining Xie, and Kaiming He.
\newblock An empirical study of training self-supervised vision transformers.
\newblock In \emph{Proceedings of the IEEE/CVF international conference on
  computer vision}, pages 9640--9649, 2021.

\bibitem[Cui et~al.(2019)Cui, Jia, Lin, Song, and Belongie]{cui2019class}
Yin Cui, Menglin Jia, Tsung-Yi Lin, Yang Song, and Serge Belongie.
\newblock Class-balanced loss based on effective number of samples.
\newblock In \emph{Proceedings of the IEEE/CVF conference on computer vision
  and pattern recognition}, pages 9268--9277, 2019.

\bibitem[Deng et~al.(2009)Deng, Dong, Socher, Li, Li, and
  Fei-Fei]{deng2009imagenet}
Jia Deng, Wei Dong, Richard Socher, Li-Jia Li, Kai Li, and Li Fei-Fei.
\newblock Imagenet: A large-scale hierarchical image database.
\newblock In \emph{2009 IEEE conference on computer vision and pattern
  recognition}, pages 248--255. Ieee, 2009.

\bibitem[Gal et~al.(2017)Gal, Islam, and Ghahramani]{gal2017deep}
Yarin Gal, Riashat Islam, and Zoubin Ghahramani.
\newblock Deep bayesian active learning with image data.
\newblock In \emph{International conference on machine learning}, pages
  1183--1192. PMLR, 2017.

\bibitem[Gissin and Shalev-Shwartz(2019)]{gissin2019discriminative}
Daniel Gissin and Shai Shalev-Shwartz.
\newblock Discriminative active learning.
\newblock \emph{arXiv preprint arXiv:1907.06347}, 2019.

\bibitem[Greer et~al.(2024)Greer, Antoniussen, Andersen, M{\o}gelmose, and
  Trivedi]{greer2024and}
Ross Greer, Bj{\o}rk Antoniussen, Mathias~V Andersen, Andreas M{\o}gelmose, and
  Mohan~M Trivedi.
\newblock The why, when, and how to use active learning in large-data-driven 3d
  object detection for safe autonomous driving: An empirical exploration.
\newblock \emph{arXiv preprint arXiv:2401.16634}, 2024.

\bibitem[Grill et~al.(2020)Grill, Strub, Altch{\'e}, Tallec, Richemond,
  Buchatskaya, Doersch, Avila~Pires, Guo, Gheshlaghi~Azar,
  et~al.]{grill2020bootstrap}
Jean-Bastien Grill, Florian Strub, Florent Altch{\'e}, Corentin Tallec, Pierre
  Richemond, Elena Buchatskaya, Carl Doersch, Bernardo Avila~Pires, Zhaohan
  Guo, Mohammad Gheshlaghi~Azar, et~al.
\newblock Bootstrap your own latent-a new approach to self-supervised learning.
\newblock \emph{Advances in neural information processing systems},
  33:\penalty0 21271--21284, 2020.

\bibitem[Hacohen et~al.(2022)Hacohen, Dekel, and Weinshall]{hacohen2022active}
Guy Hacohen, Avihu Dekel, and Daphna Weinshall.
\newblock Active learning on a budget: Opposite strategies suit high and low
  budgets.
\newblock \emph{arXiv preprint arXiv:2202.02794}, 2022.

\bibitem[He et~al.(2016)He, Zhang, Ren, and Sun]{he2016deep}
Kaiming He, Xiangyu Zhang, Shaoqing Ren, and Jian Sun.
\newblock Deep residual learning for image recognition.
\newblock In \emph{Proceedings of the IEEE conference on computer vision and
  pattern recognition}, pages 770--778, 2016.

\bibitem[Hu et~al.(2010)Hu, Mac~Namee, and Delany]{hu2010off}
Rong Hu, Brian Mac~Namee, and Sarah~Jane Delany.
\newblock Off to a good start: Using clustering to select the initial training
  set in active learning.
\newblock In \emph{FLAIRS}, 2010.

\bibitem[Janson(1986)]{janson1986random}
Svante Janson.
\newblock Random coverings in several dimensions.
\newblock 1986.

\bibitem[Jin et~al.(2022)Jin, Yuan, Qiao, and Song]{jin2022one}
Qiuye Jin, Mingzhi Yuan, Qin Qiao, and Zhijian Song.
\newblock One-shot active learning for image segmentation via contrastive
  learning and diversity-based sampling.
\newblock \emph{Knowledge-Based Systems}, 241:\penalty0 108278, 2022.

\bibitem[Kee et~al.(2018)Kee, Del~Castillo, and Runger]{kee2018query}
Seho Kee, Enrique Del~Castillo, and George Runger.
\newblock Query-by-committee improvement with diversity and density in batch
  active learning.
\newblock \emph{Information Sciences}, 454:\penalty0 401--418, 2018.

\bibitem[Krizhevsky et~al.(2009)Krizhevsky, Hinton,
  et~al.]{krizhevsky2009learning}
Alex Krizhevsky, Geoffrey Hinton, et~al.
\newblock Learning multiple layers of features from tiny images, 2009.

\bibitem[Lewis(1995)]{lewis1995sequential}
David~D Lewis.
\newblock A sequential algorithm for training text classifiers: Corrigendum and
  additional data.
\newblock In \emph{Acm Sigir Forum}, pages 13--19. ACM New York, NY, USA, 1995.

\bibitem[Liang et~al.(2024)Liang, Qiang, Ma, Wan, and Liang]{liang2024semantic}
Hailun Liang, Sunyuan Qiang, Hui Ma, Jun Wan, and Yanyan Liang.
\newblock Semantic segmentation active learning with scene coverage coreset.
\newblock In \emph{Chinese Conference on Biometric Recognition}, pages
  238--247. Springer, 2024.

\bibitem[Lin et~al.(2024)Lin, Liang, Deng, Cai, Jiang, Li, Jia, and
  Xu]{lin2024exploring}
Jinpeng Lin, Zhihao Liang, Shengheng Deng, Lile Cai, Tao Jiang, Tianrui Li, Kui
  Jia, and Xun Xu.
\newblock Exploring diversity-based active learning for 3d object detection in
  autonomous driving.
\newblock \emph{IEEE Transactions on Intelligent Transportation Systems}, 2024.

\bibitem[Maalouf et~al.(2022)Maalouf, Eini, Mussay, Feldman, and
  Osadchy]{maalouf2022unified}
Alaa Maalouf, Gilad Eini, Ben Mussay, Dan Feldman, and Margarita Osadchy.
\newblock A unified approach to coreset learning.
\newblock \emph{IEEE Transactions on Neural Networks and Learning Systems},
  2022.

\bibitem[Mishal and Weinshall(2024)]{mishal2024dcom}
Inbal Mishal and Daphna Weinshall.
\newblock Dcom: Active learning for all learners.
\newblock \emph{arXiv preprint arXiv:2407.01804}, 2024.

\bibitem[Mittal et~al.(2025)Mittal, Niemeijer, {\c{C}}i{\c{c}}ek, Tatarchenko,
  Ehrhardt, Sch{\"a}fer, Handels, and Brox]{mittal2025realistic}
Sudhanshu Mittal, Joshua Niemeijer, {\"O}zg{\"u}n {\c{C}}i{\c{c}}ek, Maxim
  Tatarchenko, Jan Ehrhardt, J{\"o}rg~P Sch{\"a}fer, Heinz Handels, and Thomas
  Brox.
\newblock Realistic evaluation of deep active learning for image classification
  and semantic segmentation.
\newblock \emph{International Journal of Computer Vision}, pages 1--23, 2025.

\bibitem[Munjal et~al.(2022)Munjal, Hayat, Hayat, Sourati, and
  Khan]{munjal2022towards}
Prateek Munjal, Nasir Hayat, Munawar Hayat, Jamshid Sourati, and Shadab Khan.
\newblock Towards robust and reproducible active learning using neural
  networks.
\newblock In \emph{Proceedings of the IEEE/CVF Conference on Computer Vision
  and Pattern Recognition}, pages 223--232, 2022.

\bibitem[Nguyen et~al.(2022)Nguyen, Shaker, and
  H{\"u}llermeier]{nguyen2022measure}
Vu-Linh Nguyen, Mohammad~Hossein Shaker, and Eyke H{\"u}llermeier.
\newblock How to measure uncertainty in uncertainty sampling for active
  learning.
\newblock \emph{Machine Learning}, 111\penalty0 (1):\penalty0 89--122, 2022.

\bibitem[Ross(2014)]{ross2014introduction}
Sheldon~M Ross.
\newblock \emph{Introduction to probability models}.
\newblock Academic press, 2014.

\bibitem[Scheffer et~al.(2001)Scheffer, Decomain, and
  Wrobel]{scheffer2001active}
Tobias Scheffer, Christian Decomain, and Stefan Wrobel.
\newblock Active hidden markov models for information extraction.
\newblock In \emph{International symposium on intelligent data analysis}, pages
  309--318. Springer, 2001.

\bibitem[Sener and Savarese(2017)]{sener2017active}
Ozan Sener and Silvio Savarese.
\newblock Active learning for convolutional neural networks: A core-set
  approach.
\newblock \emph{arXiv preprint arXiv:1708.00489}, 2017.

\bibitem[Settles(2009)]{settles2009active}
Burr Settles.
\newblock Active learning literature survey.
\newblock 2009.

\bibitem[Van~Gansbeke et~al.(2020)Van~Gansbeke, Vandenhende, Georgoulis,
  Proesmans, and Van~Gool]{van2020scan}
Wouter Van~Gansbeke, Simon Vandenhende, Stamatios Georgoulis, Marc Proesmans,
  and Luc Van~Gool.
\newblock Scan: Learning to classify images without labels.
\newblock In \emph{European conference on computer vision}, pages 268--285.
  Springer, 2020.

\bibitem[Voevodski et~al.(2012)Voevodski, Balcan, R{\"o}glin, Teng, and
  Xia]{voevodski2012active}
Konstantin Voevodski, Maria-Florina Balcan, Heiko R{\"o}glin, Shang-Hua Teng,
  and Yu Xia.
\newblock Active clustering of biological sequences.
\newblock \emph{The Journal of Machine Learning Research}, 13\penalty0
  (1):\penalty0 203--225, 2012.

\bibitem[Wang et~al.(2024)Wang, Jin, Li, Liu, Wang, and
  Song]{wang2024comprehensive}
Haoran Wang, Qiuye Jin, Shiman Li, Siyu Liu, Manning Wang, and Zhijian Song.
\newblock A comprehensive survey on deep active learning in medical image
  analysis.
\newblock \emph{Medical Image Analysis}, page 103201, 2024.

\bibitem[Wu et~al.(2023)Wu, Kang, Llambias, Ghazi, and Nielsen]{wu2023active}
Ji Wu, Zhongfeng Kang, Sebastian~N{\o}rgaard Llambias, Mostafa~Mehdipour Ghazi,
  and Mads Nielsen.
\newblock Active transfer learning for 3d hippocampus segmentation.
\newblock In \emph{Workshop on Medical Image Learning with Limited and Noisy
  Data}, pages 224--234. Springer, 2023.

\bibitem[Xu et~al.(2003)Xu, Yu, Tresp, Xu, and Wang]{xu2003representative}
Zhao Xu, Kai Yu, Volker Tresp, Xiaowei Xu, and Jizhi Wang.
\newblock Representative sampling for text classification using support vector
  machines.
\newblock In \emph{Advances in Information Retrieval: 25th European Conference
  on IR Research, ECIR 2003, Pisa, Italy, April 14--16, 2003. Proceedings 25},
  pages 393--407. Springer, 2003.

\bibitem[Yehuda et~al.(2022)Yehuda, Dekel, Hacohen, and
  Weinshall]{yehuda2022active}
Ofer Yehuda, Avihu Dekel, Guy Hacohen, and Daphna Weinshall.
\newblock Active learning through a covering lens.
\newblock \emph{Advances in Neural Information Processing Systems},
  35:\penalty0 22354--22367, 2022.

\bibitem[Zakariah and Almazyad(2023)]{zakariah2023anomaly}
Mohammed Zakariah and Abdulaziz~S Almazyad.
\newblock Anomaly detection for iot systems using active learning.
\newblock \emph{Applied Sciences}, 13\penalty0 (21):\penalty0 12029, 2023.

\bibitem[Zhang et~al.(2023)Zhang, Chen, Canal, Mussmann, Das, Bhatt, Zhu,
  Bilmes, Du, Jamieson, et~al.]{zhang2023labelbench}
Jifan Zhang, Yifang Chen, Gregory Canal, Stephen Mussmann, Arnav~M Das,
  Gantavya Bhatt, Yinglun Zhu, Jeffrey Bilmes, Simon~Shaolei Du, Kevin
  Jamieson, et~al.
\newblock Labelbench: A comprehensive framework for benchmarking adaptive
  label-efficient learning.
\newblock \emph{arXiv preprint arXiv:2306.09910}, 2023.

\end{thebibliography}
}

\clearpage
\appendix
\section{Supplementary Materials for PALM}

\subsection{Related Work}

Active learning (AL) has been widely explored as a means to reduce annotation costs by querying the most informative samples. AL strategies are commonly categorized by their core selection criteria, including uncertainty, diversity, representation, and hybrid approaches. Uncertainty-based methods prioritize samples where the model exhibits the lowest confidence. A classical approach is uncertainty sampling \cite{lewis1995sequential}, which selects data points with low predicted class confidence. Margin sampling \cite{scheffer2001active} targets instances where the difference between the top two predicted class probabilities is small, indicating ambiguity. Entropy-based sampling \cite{settles2009active} captures total predictive uncertainty by selecting samples with high entropy in the output distribution.

Diversity-based methods seek to avoid redundancy by selecting a set of samples that spans the data distribution. These techniques often rely on geometric or statistical distance metrics. For example, the k-Center Greedy algorithm \cite{sener2017active} minimizes the maximum distance between selected points and the remaining pool. BADGE \cite{ash2019deep} combines uncertainty and diversity by clustering in gradient space using a k-means++ scheme. Diversity plays a crucial role in early-stage selection to ensure broader coverage of the input space. Moreover, representation-based methods utilize structure in the feature space to guide sampling. These often rely on clustering or geometric criteria to identify representative or central points. Examples include k-means-based sampling \cite{xu2003representative}, medoid selection \cite{aghaee2016active}, and median-based heuristics \cite{voevodski2012active}. TypiClust \cite{hacohen2022active} extends this idea by combining sample typicality and cluster centrality, favoring samples that are both generalizable and diverse.

Hybrid methods integrate multiple selection criteria, often combining uncertainty with diversity. DBAL \cite{gal2017deep} leverages Bayesian dropout to estimate uncertainty, while promoting diversity among queried samples. These methods are particularly valuable in deep neural networks, where relying solely on uncertainty can lead to redundant or misleading selections. To facilitate the empirical comparison of AL strategies, LabelBench \cite{zhang2023labelbench} was proposed as a modular and extensible benchmarking suite. It enables evaluation of AL, semi-supervised learning, and transfer learning under consistent conditions, including different model architectures and labeling budgets. LabelBench places strong emphasis on reproducibility and explores the synergy between AL and pretrained models, particularly vision transformers. Its findings suggest that combining AL with SSL can yield notable improvements in label efficiency.

While LabelBench provides a valuable empirical benchmarking framework, our work contributes a complementary modeling perspective. PALM introduces a predictive and interpretable parametric model for characterizing AL behavior. By estimating three key descriptors, i.e., initial performance, growth rate, and asymptotic accuracy, PALM enables quantitative comparison of AL methods and forecasting of future performance based on partial observations. Although both approaches evaluate common AL strategies, LabelBench emphasizes empirical performance across tasks and architectures, while PALM focuses on modeling and interpretability of AL dynamics.

\subsection{PALM Proofs and Corollary}

In this section, we provide the proofs and a corollary corresponding to the methods described in the main text.

\subsubsection{Definition 1: Coverage Probability}

\textit{Proof.}
By the complement rule in probability theory, the probability of an event occurring is equal to one minus the probability of its complement. Let $A$ represent the event that a point $x$ is covered by at least one object, and let $A^c$ represent the complement event, where $x$ is not covered by any object. According to the complement rule, we have:
\begin{equation}
    P(A) + P(A^c) = 1.
\end{equation}
Substituting $P(A) = P_{\text{C}}$ and $P(A^c) = P_{\text{NC}}$, the equation becomes:
\begin{equation}
    P_{\text{C}} + P_{\text{NC}} = 1,
\end{equation}
which completes the proof.

\subsubsection{Definition 2: Coverage Probability with s Independent Objects}

\textit{Proof.}
Let $p$ represent the probability that a single randomly placed object covers point $x$. The probability that a single object does not cover $x$ is $1 - p$. Now, consider $s$ objects placed independently in the space $\mathbb{X}$. Since the objects are independent, the probability that none of them covers $x$ is the product of their individual non-coverage probabilities:
\begin{equation}
    P_{\text{UC}} = (1 - p)^s.
\end{equation}
Thus, the probability that $x$ is covered by at least one object is:
\begin{equation}
    P_{\text{C}} = 1 - (1 - p)^s.
\end{equation}
This completes the proof.

\subsubsection{Corollary 2: Asymptotic Behavior of Accuracy as a Function of Coverage Probability}
\label{corollary2}

The test generalization accuracy function is given by:
\begin{equation}
    \text{A} = \text{A}_{C} \, \left(1 - (1 - \delta)^B\right) + \text{A}_{UC} \, (1 - \delta)^B.
\end{equation}
This function exhibits the following asymptotic behaviors:

Case 1. No Labeled Samples ($B = 0$): When no labeled samples are available, the coverage fraction is:
\begin{equation}
    P_{\text{C}} = 1 - (1 - \delta)^0 = 0.
\end{equation}
Substituting this into the accuracy function gives $\text{A} = \text{A}_{\text{UC}}$, indicating that without labeled data, the model's accuracy depends on its performance in the uncovered regions.

Case 2. Infinite Labeled Samples ($B \to \infty$): As the number of labeled samples increases, the coverage probability approaches one:
\begin{equation}
    \lim_{B \to \infty} (1 - \delta)^B = 0.
\end{equation}
Substituting this into the accuracy function yields:
\begin{equation}
    \lim_{B \to \infty} \text{A} = \text{A}_{\text{C}},
\end{equation}
which implies that with full coverage, the model achieves its maximum accuracy in the covered regions.

Case 3. Small $\delta$ Approximation for Large $B$: For small values of $\delta$, the coverage term $(1 - \delta)^B$ can be approximated using the first-order Taylor expansion:
\begin{equation}
    (1 - \delta)^B \approx e^{-B\delta},
\end{equation}
since $\lim_{x \to 0^+} (1 - x) \approx e^{-x}$. Thus, for sufficiently large $B$, the accuracy function approximates:
\begin{equation}
    \text{A} \approx \text{A}_{\text{C}} \, \left(1 - e^{-B\delta}\right) + \text{A}_{\text{UC}} \, e^{-B\delta},
\end{equation}
which shows that the accuracy converges exponentially towards $\text{A}_{\text{C}}$, with the rate of convergence governed by $\delta$.

\subsubsection{Definition 6: Generalized Accuracy as a Function of Coverage with Exponential Adjustment}

\textit{Proof.}
We aim to derive the generalized accuracy function, which incorporates the parameters $\alpha$ and $\beta$. The function is defined as:
\begin{equation}
    \text{A} = \text{A}_{\max} \left(1 - (1 - \delta)^{(B + \alpha)^\beta} \right),
\end{equation}
where $\text{A}_{\max}$ is the maximum achievable accuracy under full coverage, $B$ is the cumulative number of labeled samples, $\delta$ represents the expected fraction of the space covered by a single labeled sample, $\alpha$ accounts for initial learning effects and prior knowledge, allowing non-zero accuracy when $B = 0$, and $\beta$ controls the scaling of accuracy growth as $B$ increases.

We start with the test accuracy function given by:
\begin{equation}
    \text{A} = \text{A}_{\text{C}} \, \left(1 - (1 - \delta)^B\right) + \text{A}_{\text{UC}} \, (1 - \delta)^B.
\end{equation}
Rearranging the terms:
\begin{equation}
    \text{A} = \text{A}_{\text{C}} - (\text{A}_{\text{C}} - \text{A}_{\text{UC}}) (1 - \delta)^B.
\end{equation}
Assuming that $\text{A}_{\max} = \text{A}_{\text{C}}$, we rewrite the expression as:
\begin{equation}
    \text{A} = \text{A}_{\max} - (\text{A}_{\max} - \text{A}_{\text{UC}}) (1 - \delta)^B.
\end{equation}
As $B$ increases, the second term vanishes, since $(1 - \delta)^B \to 0$, which ensures that $\text{A} \to \text{A}_{\max}$, as expected under full coverage. To generalize this formulation and account for variations in early learning dynamics and growth rates, we replace $B$ with the adjusted term $(B + \alpha)^\beta$, where $\alpha > 0$ allows the model to exhibit non-zero accuracy even when $B = 0$, representing prior knowledge or inherent generalization, and $\beta > 0$ modulates the rate of accuracy increase with $B$. Substituting this adjustment, the generalized accuracy function becomes:
\begin{equation}
    \text{A} = \text{A}_{\max} - (\text{A}_{\max} - \text{A}_{\text{UC}}) (1 - \delta)^{(B + \alpha)^\beta}.
\end{equation}
Finally, assuming that the uncovered regions contribute negligible accuracy ($\text{A}_{\text{UC}} \approx 0$), the expression simplifies to:
\begin{equation}
    \text{A} = \text{A}_{\max} \left(1 - (1 - \delta)^{(B + \alpha)^\beta} \right).
\end{equation}
Thus, the generalized accuracy function models the influence of labeled sample coverage, prior knowledge, and learning dynamics on active learning performance.

\subsubsection{Corollary 3: Asymptotic Behavior of Generalized Accuracy with Exponential Adjustment}
\label{corollary3}

The generalized test accuracy function exhibits the following asymptotic behavior:

Case 1. No Labeled Samples ($B = 0$): When no labeled samples are available, the accuracy simplifies to:
\begin{equation}
    \text{A} = \text{A}_{\max} \left(1 - (1 - \delta)^{\alpha^\beta} \right),
\end{equation}
which leads to two characteristic scenarios:
\begin{itemize}
    \item If $\alpha > 0$, then $\text{A} > 0$, indicating that the model achieves non-zero accuracy even without labeled data. This reflects the model's ability to generalize from uncovered regions or prior knowledge.
    \item If $\alpha = 0$, we recover the classical case where no coverage implies zero accuracy, i.e., $\text{A} = 0$.
\end{itemize}

Case 2. Infinite Labeled Samples ($B \to \infty$): As the number of labeled samples grows to infinity:
\begin{equation}
    \lim_{B \to \infty} (B + \alpha)^\beta = \infty.
\end{equation}
Since $0 < 1 - \delta < 1$, we have:
\begin{equation}
    \lim_{B \to \infty} (1 - \delta)^{(B + \alpha)^\beta} = 0.
\end{equation}
Therefore, the accuracy converges to its theoretical maximum:
\begin{equation}
    \lim_{B \to \infty} \text{A} = \text{A}_{\max}.
\end{equation}
This confirms that, with full coverage, the model achieves optimal performance.

Case 3. Small $\delta$ Approximation for Large $B$: For small $\delta$, the coverage term can be approximated using the first-order Taylor expansion:
\begin{equation}
    (1 - \delta)^{(B + \alpha)^\beta} \approx e^{-(B + \alpha)^\beta \delta}.
\end{equation}
Thus, for sufficiently large $B$, the accuracy function approximates:
\begin{equation}
    \text{A} \approx \text{A}_{\max} \left(1 - e^{-(B + \alpha)^\beta \delta} \right),
\end{equation}
which shows that the accuracy converges exponentially toward $\text{A}_{\max}$, with a rate of convergence determined by $\alpha$, $\beta$, and $\delta$.

\subsubsection{Lemma 1: Parameter Estimation for Learning Dynamics in Active Learning Without Normalization}
\label{lemmma1}

Let $B$ denote the total number of labeled samples collected during an active learning (AL) process. Given the observed accuracy values from the AL process, we aim to estimate the parameters $\text{A}_{\max}$, $\delta$, $\alpha$, and $\beta$ in the model:
\begin{equation}
    \text{A} = \text{A}_{\max} \left(1 - (1 - \delta)^{(B + \alpha)^\beta} \right).
\end{equation}
These parameters can be empirically estimated from accuracy measurements collected over multiple AL iterations without the need to normalize $B$.

Suppose accuracy is observed for at least four different cumulative budgets $B_1, B_2, B_3, B_4$, with corresponding accuracies $\text{A}_1$, $\text{A}_2$, $\text{A}_3$, and $\text{A}_4$. The parameters $\text{A}_{\max}$, $\delta$, $\alpha$, and $\beta$ can then be estimated by solving the following system of equations using nonlinear regression techniques:
\begin{equation}
    \text{A}_i = \text{A}_{\max} \left(1 - (1 - \delta)^{(B_i + \alpha)^\beta} \right), \quad i = 1, 2, 3, 4.
\end{equation}

The complexity of estimating the parameters primarily depends on evaluating the exponentiation term $(1 - \delta)^{(B + \alpha)^\beta}$. Using optimized exponentiation algorithms, the complexity is approximately $\mathcal{O}(\log(B))$. However, in naive implementations, the complexity can approach $\mathcal{O}(B^\beta)$, especially for large $B$. For large values of $B$, the following challenges arise:
\begin{itemize}
    \item Computational Overhead: The term $(B + \alpha)^\beta$ grows rapidly, increasing computation time.
    \item Numerical Instability: Large exponents may lead to floating-point precision errors.
    \item Diminishing Accuracy Gains: As $B$ increases, the marginal contribution of additional labeled samples decreases due to saturation effects.
\end{itemize}
To mitigate these issues, normalizing $B$ by the mean budget per iteration ($b$) reduces the computational cost from $\mathcal{O}(B^\beta)$ to $\mathcal{O}((B/b)^\beta)$, improves numerical stability during exponentiation, and ensures smoother convergence behavior of the accuracy function. The normalized generalized accuracy function is then given by:
\begin{equation}
    \text{A} = \text{A}_{\max} \left(1 - (1 - \delta)^{\left( \frac{B}{b} + \alpha \right)^\beta} \right),
\end{equation}
where normalization aligns the function with the number of AL iterations rather than the absolute number of labeled samples.

\subsubsection{Theorem 1: Comparing Two Active Learning Methods Using the Normalized Accuracy Function}
\label{theorem1}

\textit{Proof.}
Consider two active learning methods 1 and 2, with normalized accuracy functions defined as:
\begin{align}
\text{A}_1 &= \text{A}_{\max, 1} \left(1 - (1 - \delta_1)^{\left( \frac{B}{b_1} + \alpha_1 \right)^{\beta_1}} \right), \\
\text{A}_2 &= \text{A}_{\max, 2} \left(1 - (1 - \delta_2)^{\left( \frac{B}{b_2} + \alpha_2 \right)^{\beta_2}} \right).
\end{align}
To compare their performance for a given budget $B$, we define the ratio of accuracies. Method 1 outperforms Method 2 at budget $B$ if $\text{A}_1 / \text{A}_2 > 1$.

\textbf{Full Coverage Limit ($B \to \infty$).} As the number of labeled samples approaches infinity:
\begin{equation}
\lim_{B \to \infty} \text{A}_1 = \text{A}_{\max, 1}, \quad \lim_{B \to \infty} \text{A}_2 = \text{A}_{\max, 2}.
\end{equation}
Thus, in the limit of infinite budget, the method with the higher $\text{A}_{\max}$ dominates:
\begin{equation}
\text{A}_{\max, 1} > \text{A}_{\max, 2} \quad \Rightarrow \quad \lim_{B \to \infty} \frac{\text{A}_1}{\text{A}_2} > 1.
\end{equation}

\textbf{Early-Stage Learning (small $B$).} For small budget $B$, applying the first-order Taylor approximation results in:
\begin{equation}
(1 - \delta)^x \approx e^{-x\delta}.
\end{equation}
Therefore, we can approximate the accuracy functions as:
\begin{align}
\text{A}_1 &\approx \text{A}_{\max, 1} \left(1 - e^{-\delta_1 \left( \frac{B}{b_1} + \alpha_1 \right)^{\beta_1}} \right), \\
\text{A}_2 &\approx \text{A}_{\max, 2} \left(1 - e^{-\delta_2 \left( \frac{B}{b_2} + \alpha_2 \right)^{\beta_2}} \right).
\end{align}
In this regime, faster accuracy growth occurs for the method with larger $\delta$, higher $\alpha$, smaller $b$, and larger $\beta$.

\textbf{General Comparison Criterion.} To compare the accuracy growth rates, differentiate the accuracy functions with respect to $B$. Method 1 improves faster than Method 2 if:
\begin{equation}
\frac{d\, \text{A}_1}{dB} > \frac{d\, \text{A}_2}{dB}.
\end{equation}
This condition holds when:
\begin{equation}
\delta_1 \left( \frac{1}{b_1} + \frac{\alpha_1}{B} \right)^{\beta_1} > \delta_2 \left( \frac{1}{b_2} + \frac{\alpha_2}{B} \right)^{\beta_2}.
\end{equation}

In summary, Method 1 outperforms Method 2 when it exhibits higher coverage efficiency ($\delta$), smaller batch size ($b$), greater initial accuracy boost ($\alpha$), or faster accuracy scaling ($\beta$). Additionally, the asymptotic accuracy $\text{A}_{\max}$ determines long-term dominance as $B$ increases. Together, these parameters provide a comprehensive framework for quantitatively comparing active learning strategies across different budget regimes.

\newpage
\subsection{Quantitative Results}

\begin{table*}[ht]
\centering
\caption{PALM parameter estimates for CIFAR-10 without pretrained embeddings, evaluated across various AL strategies and different numbers of labeled points used for curve fitting based on the mean values from 5 repetitions. The table reports the maximum achievable accuracy ($\text{A}_{\max}$), coverage efficiency ($\delta$), early-stage performance offset ($\alpha$), and scalability ($\beta$). In the absence of pretrained embeddings, methods show slower learning dynamics and lower $\delta$ values, with $\alpha$ increasing over time, indicating delayed accuracy gains. TypiClust demonstrates relatively higher $\delta$ values throughout, reflecting strong sample efficiency. In contrast, methods like Margin and Entropy show increasing $\alpha$ and $\beta$, indicating slower convergence in later stages.}
\label{tab:cifar10_no_ssl_params1}
\resizebox{\textwidth}{!}{
\begin{tabular}{lcccccccccccccccccccccccccccc}
\toprule
\multirow{2}{*}{\textbf{AL Method}} 
& \multicolumn{4}{c}{\textbf{6 Points}} 
& \multicolumn{4}{c}{\textbf{10 Points}} 
& \multicolumn{4}{c}{\textbf{20 Points}} 
& \multicolumn{4}{c}{\textbf{50 Points}} 
& \multicolumn{4}{c}{\textbf{100 Points}} 
& \multicolumn{4}{c}{\textbf{500 Points}} 
& \multicolumn{4}{c}{\textbf{1000 Points}} \\
\cmidrule(lr){2-5} \cmidrule(lr){6-9} \cmidrule(lr){10-13} \cmidrule(lr){14-17} \cmidrule(lr){18-21} \cmidrule(lr){22-25} \cmidrule(lr){26-29}
& $\text{A}_{\max}$ & $\delta$ & $\alpha$ & $\beta$
& $\text{A}_{\max}$ & $\delta$ & $\alpha$ & $\beta$
& $\text{A}_{\max}$ & $\delta$ & $\alpha$ & $\beta$
& $\text{A}_{\max}$ & $\delta$ & $\alpha$ & $\beta$
& $\text{A}_{\max}$ & $\delta$ & $\alpha$ & $\beta$
& $\text{A}_{\max}$ & $\delta$ & $\alpha$ & $\beta$
& $\text{A}_{\max}$ & $\delta$ & $\alpha$ & $\beta$ \\
\midrule
Random 
& 32.6 & 0.373 & 2.519 & 0.688
& 100 & 0.168 & 1.611 & 0.303
& 100 & 0.144 & 2.466 & 0.365
& 100 & 0.139 & 2.745 & 0.376
& 100 & 0.124 & 3.696 & 0.409
& 93.8 & 0.116 & 4.951 & 0.446
& 90.7 & 0.108 & 6.263 & 0.475 \\
Uncertainty 
& 100 & 0.105 & 1.714 & 0.467
& 36.3 & 0.138 & 3.365 & 0.932
& 49.0 & 0.147 & 2.716 & 0.701
& 65.2 & 0.161 & 1.624 & 0.510
& 91.5 & 0.135 & 1.092 & 0.416
& 99.6 & 0.102 & 2.614 & 0.456
& 93.7 & 0.090 & 4.644 & 0.506 \\
Margin 
& 39.1 & 0.457 & 1.289 & 0.390
& 100 & 0.167 & 1.672 & 0.316
& 100 & 0.150 & 2.227 & 0.360
& 100 & 0.142 & 2.626 & 0.377
& 100 & 0.127 & 3.587 & 0.409
& 93.6 & 0.097 & 7.353 & 0.497
& 92.6 & 0.094 & 7.812 & 0.507 \\
Entropy 
& 56.7 & 0.317 & 0.355 & 0.282
& 100 & 0.176 & 0.397 & 0.264
& 100 & 0.165 & 0.611 & 0.304
& 100 & 0.152 & 0.935 & 0.333
& 100 & 0.126 & 2.204 & 0.389
& 99.4 & 0.083 & 7.303 & 0.490
& 93.3 & 0.070 & 10.00 & 0.546 \\
TypiClust 
& 35.4 & 0.489 & 1.274 & 0.617
& 42.3 & 0.474 & 0.858 & 0.412
& 52.3 & 0.410 & 0.618 & 0.308
& 100 & 0.180 & 1.726 & 0.307
& 100 & 0.165 & 2.451 & 0.333
& - & - & - & -
& - & - & - & - \\
\bottomrule
\end{tabular}
}
\end{table*}

\begin{table*}[ht]
\centering
\caption{PALM parameter estimates for CIFAR-10 without pretrained embeddings, evaluated across various AL strategies and different numbers of labeled points used for curve fitting based on the minimum values from 5 repetitions. The table reports the maximum achievable accuracy ($\text{A}_{\max}$), coverage efficiency ($\delta$), early-stage performance offset ($\alpha$), and scalability ($\beta$). In the absence of pretrained embeddings, methods show slower learning dynamics and lower $\delta$ values, with $\alpha$ increasing over time, indicating delayed accuracy gains. TypiClust demonstrates relatively higher $\delta$ values throughout, reflecting strong sample efficiency. In contrast, methods like Margin and Entropy show increasing $\alpha$ and $\beta$, indicating slower convergence in later stages.}
\label{tab:cifar10_no_ssl_params2}
\resizebox{\textwidth}{!}{
\begin{tabular}{lcccccccccccccccccccccccccccc}
\toprule
\multirow{2}{*}{\textbf{AL Method}} 
& \multicolumn{4}{c}{\textbf{6 Points}} 
& \multicolumn{4}{c}{\textbf{10 Points}} 
& \multicolumn{4}{c}{\textbf{20 Points}} 
& \multicolumn{4}{c}{\textbf{50 Points}} 
& \multicolumn{4}{c}{\textbf{100 Points}} 
& \multicolumn{4}{c}{\textbf{500 Points}} 
& \multicolumn{4}{c}{\textbf{1000 Points}} \\
\cmidrule(lr){2-5} \cmidrule(lr){6-9} \cmidrule(lr){10-13} \cmidrule(lr){14-17} \cmidrule(lr){18-21} \cmidrule(lr){22-25} \cmidrule(lr){26-29}
& $\text{A}_{\max}$ & $\delta$ & $\alpha$ & $\beta$
& $\text{A}_{\max}$ & $\delta$ & $\alpha$ & $\beta$
& $\text{A}_{\max}$ & $\delta$ & $\alpha$ & $\beta$
& $\text{A}_{\max}$ & $\delta$ & $\alpha$ & $\beta$
& $\text{A}_{\max}$ & $\delta$ & $\alpha$ & $\beta$
& $\text{A}_{\max}$ & $\delta$ & $\alpha$ & $\beta$
& $\text{A}_{\max}$ & $\delta$ & $\alpha$ & $\beta$ \\
\midrule
Random
& 100 & 0.144 & 2.369 & 0.319
& 100 & 0.085 & 4.819 & 0.534
 & 100 & 0.093   & 4.654 & 0.493
& 87.1 & 0.143 & 2.824 & 0.407
& 100 & 0.114 & 3.635 & 0.418
& 93.4  & 0.107 & 4.930  & 0.458
& 90.0  & 0.098 & 6.350  & 0.490 \\
Uncertainty
& 73.7  & 0.000 & 38.91 & 4.559
& 57.9  & 0.000 & 56.46 & 4.146
& 50.7  & 0.148   & 1.790 & 0.658
& 57.7  & 0.154   & 1.355 & 0.571
& 81.6  & 0.138   & 0.732 & 0.442
& 100  & 0.095  & 2.140 & 0.462
& 93.7  & 0.083  & 4.104 & 0.512 \\
Margin
& 70.8  & 0.000   & 39.14 & 4.522
& 100 & 0.188   & 0.446  & 0.246
& 100 & 0.164   & 0.955  & 0.319
& 100 & 0.142   & 1.733  & 0.369
& 100 & 0.125   & 2.669  & 0.406
& 97.3  & 0.104   & 4.834  & 0.460
& 92.8  & 0.093   & 6.879  & 0.502 \\
Entropy
& 72.5  & 0.000   & 40.98 & 4.510
& 55.6  & 0.000   & 53.58 & 4.138 
& 100 & 0.146   & 0.380  & 0.327
& 100 & 0.136   & 0.599  & 0.356
& 100 & 0.122   & 1.093  & 0.389
& 100 & 0.077   & 6.406  & 0.495
& 93.2  & 0.063   & 10.31 & 0.559 \\
TypiClust
& 63.5  & 0.000   & 35.52& 4.594
& 100 & 0.197   & 0.508  & 0.262
& 100 & 0.203   & 0.409  & 0.242
& 48.2  & 0.343   & 1.120  & 0.425
& 49.9  & 0.353   & 0.878  & 0.386
& - & - & - & -
& - & - & - & - \\
\bottomrule
\end{tabular}
}
\end{table*}

\begin{table*}[ht]
\centering
\caption{PALM parameter estimates for CIFAR-10 without pretrained embeddings, evaluated across various AL strategies and different numbers of labeled points used for curve fitting based on the maximum values from 5 repetitions. The table reports the maximum achievable accuracy ($\text{A}_{\max}$), coverage efficiency ($\delta$), early-stage performance offset ($\alpha$), and scalability ($\beta$). In the absence of pretrained embeddings, methods show slower learning dynamics and lower $\delta$ values, with $\alpha$ increasing over time, indicating delayed accuracy gains. TypiClust demonstrates relatively higher $\delta$ values throughout, reflecting strong sample efficiency. In contrast, methods like Margin and Entropy show increasing $\alpha$ and $\beta$, indicating slower convergence in later stages.}
\label{tab:cifar10_no_ssl_params3}
\resizebox{\textwidth}{!}{
\begin{tabular}{lcccccccccccccccccccccccccccc}
\toprule
\multirow{2}{*}{\textbf{AL Method}} 
& \multicolumn{4}{c}{\textbf{6 Points}} 
& \multicolumn{4}{c}{\textbf{10 Points}} 
& \multicolumn{4}{c}{\textbf{20 Points}} 
& \multicolumn{4}{c}{\textbf{50 Points}} 
& \multicolumn{4}{c}{\textbf{100 Points}} 
& \multicolumn{4}{c}{\textbf{500 Points}} 
& \multicolumn{4}{c}{\textbf{1000 Points}} \\
\cmidrule(lr){2-5} \cmidrule(lr){6-9} \cmidrule(lr){10-13} \cmidrule(lr){14-17} \cmidrule(lr){18-21} \cmidrule(lr){22-25} \cmidrule(lr){26-29}
& $\text{A}_{\max}$ & $\delta$ & $\alpha$ & $\beta$
& $\text{A}_{\max}$ & $\delta$ & $\alpha$ & $\beta$
& $\text{A}_{\max}$ & $\delta$ & $\alpha$ & $\beta$
& $\text{A}_{\max}$ & $\delta$ & $\alpha$ & $\beta$
& $\text{A}_{\max}$ & $\delta$ & $\alpha$ & $\beta$
& $\text{A}_{\max}$ & $\delta$ & $\alpha$ & $\beta$
& $\text{A}_{\max}$ & $\delta$ & $\alpha$ & $\beta$ \\
\midrule
Random
& 29.6  & 0.001 & 16.87 & 5.348
& 34.8  & 0.423    & 2.032 & 0.639
& 100  & 0.180      & 1.608 & 0.301
& 100  & 0.145    & 3.133 & 0.372
& 100  & 0.126    & 4.355 & 0.412
& 100  & 0.126    & 4.355 & 0.412
& 91.6  & 0.120    & 5.799 & 0.454 \\
Uncertainty
& 100 & 0.008 & 11.58 & 1.265
& 33.1  & 0.000 & 28.16 & 4.467
& 40.4  & 0.022 & 9.550  & 1.423
& 71.9  & 0.150 & 2.545  & 0.495
& 99.5  & 0.123 & 2.053  & 0.420
& 94.8  & 0.101 & 4.184  & 0.486
& 93.8  & 0.098 & 4.613  & 0.496 \\
Margin
& 100 & 0.092   & 6.633  & 0.520
& 100 & 0.000   & 44.99 & 2.612
& 41.2  & 0.007   & 16.22 & 1.686
& 100 & 0.139   & 3.940  & 0.389
& 100 & 0.123   & 5.139  & 0.421
& 91.1  & 0.088   & 10.16 & 0.534
& 92.6  & 0.097   & 8.479  & 0.506 \\
Entropy
& 47.9  & 0.000   & 81.58 & 3.796
& 100 & 0.155   & 1.531  & 0.346
& 100 & 0.159   & 1.406  & 0.336
& 80.2  & 0.200   & 1.328  & 0.348
& 100 & 0.133   & 2.932  & 0.384
& 91.8  & 0.067   & 12.81 & 0.569
& 92.7  & 0.072   & 11.69 & 0.553 \\
TypiClust
& 33.8  & 0.222  & 2.706  & 1.290
& 36.2  & 0.456  & 1.481  & 0.749
& 48.3  & 0.481  & 0.550  & 0.324
& 100 & 0.165  & 3.205  & 0.345
& 100 & 0.142  & 4.792  & 0.388
& - & - & - & -
& - & - & - & - \\
\bottomrule
\end{tabular}
}
\end{table*}

\begin{table*}[ht]
\centering
\caption{PALM parameter estimates for CIFAR-10 using SimCLR embeddings for feature extraction, evaluated across various AL strategies and different numbers of labeled points used for curve fitting based on the mean values from 5 repetitions. The table reports the maximum achievable accuracy ($\text{A}_{\max}$), coverage efficiency ($\delta$), early-stage performance offset ($\alpha$), and scalability ($\beta$). The results highlight the acceleration of learning dynamics with pretrained embeddings, where TypiClust and Margin benefit from high $\delta$ and low $\alpha$, indicating efficient early-stage learning. Conversely, methods like Entropy and DBAL exhibit delayed improvements at small budgets but show recovery and better performance as the annotation grows.}
\label{tab:cifar10_simclr_params1}
\resizebox{\textwidth}{!}{
\begin{tabular}{lcccccccccccccccccccccccccccc}
\toprule
\multirow{2}{*}{\textbf{AL Method}} 
& \multicolumn{4}{c}{\textbf{6 Points}} 
& \multicolumn{4}{c}{\textbf{10 Points}} 
& \multicolumn{4}{c}{\textbf{20 Points}} 
& \multicolumn{4}{c}{\textbf{50 Points}} 
& \multicolumn{4}{c}{\textbf{100 Points}} 
& \multicolumn{4}{c}{\textbf{500 Points}} 
& \multicolumn{4}{c}{\textbf{1000 Points}} \\
\cmidrule(lr){2-5} \cmidrule(lr){6-9} \cmidrule(lr){10-13} \cmidrule(lr){14-17} \cmidrule(lr){18-21} \cmidrule(lr){22-25} \cmidrule(lr){26-29}
& $\text{A}_{\max}$ & $\delta$ & $\alpha$ & $\beta$
& $\text{A}_{\max}$ & $\delta$ & $\alpha$ & $\beta$
& $\text{A}_{\max}$ & $\delta$ & $\alpha$ & $\beta$
& $\text{A}_{\max}$ & $\delta$ & $\alpha$ & $\beta$
& $\text{A}_{\max}$ & $\delta$ & $\alpha$ & $\beta$
& $\text{A}_{\max}$ & $\delta$ & $\alpha$ & $\beta$
& $\text{A}_{\max}$ & $\delta$ & $\alpha$ & $\beta$ \\
\midrule
Random 
& 100 & 0.443 & 0.536 & 0.341
& 100 & 0.451 & 0.473 & 0.317
& 100 & 0.394 & 0.912 & 0.413
& 84.2 & 0.290 & 2.314 & 0.735
& 85.0 & 0.355 & 1.690 & 0.633
& 86.6 & 0.496 & 0.609 & 0.434
& 87.1 & 0.536 & 0.392 & 0.381 \\
Uncertainty 
& 100 & 0.278 & 0.123 & 0.147
& 74.1 & 0.000 & 26.35 & 4.534
& 75.2 & 0.000 & 22.50 & 4.536
& 79.8 & 0.005 & 8.341 & 1.940
& 82.7 & 0.061 & 3.390 & 1.142
& 86.6 & 0.220 & 0.708 & 0.640
& 87.6 & 0.270 & 0.397 & 0.546 \\
Margin 
& 68.1 & 0.000 & 12.57 & 5.699
& 80.1 & 0.250 & 1.853 & 0.966
& 82.5 & 0.286 & 1.596 & 0.857
& 84.3 & 0.357 & 1.133 & 0.705
& 85.5 & 0.409 & 0.819 & 0.605
& 87.5 & 0.511 & 0.326 & 0.428
& 87.9 & 0.535 & 0.241 & 0.389 \\
Entropy 
& 34.8 & 0.000 & 13.34 & 5.462
& 39.1 & 0.299 & 1.822 & 0.856
& 81.6 & 0.000 & 31.60 & 4.337
& 76.6 & 0.000 & 31.40 & 4.427
& 80.2 & 0.008 & 8.037 & 1.643
& 86.2 & 0.183 & 0.581 & 0.632
& 87.5 & 0.227 & 0.305 & 0.542 \\
TypiClust 
& 78.1 & 0.859 & 0.536 & 0.347
& 88.2 & 0.788 & 0.188 & 0.147
& 96.7 & 0.721 & 0.153 & 0.116
& 85.5 & 0.803 & 0.298 & 0.188
& 85.3 & 0.803 & 0.312 & 0.192
& - & - & - & -
& - & - & - & - \\
DBAL 
& 33.3 & 0.000 & 12.11 & 5.595
& 100 & 0.185 & 0.862 & 0.369
& 79.6 & 0.000 & 30.30 & 4.377
& 76.1 & 0.000 & 29.55 & 4.413
& 80.0 & 0.014 & 6.508 & 1.505
& 86.3 & 0.197 & 0.488 & 0.608
& 87.5 & 0.236 & 0.274 & 0.528 \\
\bottomrule
\end{tabular}
}
\end{table*}

\begin{table*}[ht]
\centering
\caption{PALM parameter estimates for CIFAR-10 using SimCLR embeddings for feature extraction, evaluated across various AL strategies and different numbers of labeled points used for curve fitting based on the minimum values from 5 repetitions. The table reports the maximum achievable accuracy ($\text{A}_{\max}$), coverage efficiency ($\delta$), early-stage performance offset ($\alpha$), and scalability ($\beta$). The results highlight the acceleration of learning dynamics with pretrained embeddings, where TypiClust and Margin benefit from high $\delta$ and low $\alpha$, indicating efficient early-stage learning. Conversely, methods like Entropy and DBAL exhibit delayed improvements at small budgets but show recovery and better performance as the annotation grows.}
\label{tab:cifar10_simclr_params2}
\resizebox{\textwidth}{!}{
\begin{tabular}{lcccccccccccccccccccccccccccc}
\toprule
\multirow{2}{*}{\textbf{AL Method}} 
& \multicolumn{4}{c}{\textbf{6 Points}} 
& \multicolumn{4}{c}{\textbf{10 Points}} 
& \multicolumn{4}{c}{\textbf{20 Points}} 
& \multicolumn{4}{c}{\textbf{50 Points}} 
& \multicolumn{4}{c}{\textbf{100 Points}} 
& \multicolumn{4}{c}{\textbf{500 Points}} 
& \multicolumn{4}{c}{\textbf{1000 Points}} \\
\cmidrule(lr){2-5} \cmidrule(lr){6-9} \cmidrule(lr){10-13} \cmidrule(lr){14-17} \cmidrule(lr){18-21} \cmidrule(lr){22-25} \cmidrule(lr){26-29}
& $\text{A}_{\max}$ & $\delta$ & $\alpha$ & $\beta$
& $\text{A}_{\max}$ & $\delta$ & $\alpha$ & $\beta$
& $\text{A}_{\max}$ & $\delta$ & $\alpha$ & $\beta$
& $\text{A}_{\max}$ & $\delta$ & $\alpha$ & $\beta$
& $\text{A}_{\max}$ & $\delta$ & $\alpha$ & $\beta$
& $\text{A}_{\max}$ & $\delta$ & $\alpha$ & $\beta$
& $\text{A}_{\max}$ & $\delta$ & $\alpha$ & $\beta$ \\
\midrule
Random
& 60.4  & 0.000  & 12.60 & 5.593
& 66.7  & 0.510  & 0.808  & 0.588
& 100 & 0.258  & 1.699  & 0.562
& 83.5  & 0.117  & 4.241  & 1.009
& 84.6  & 0.211  & 2.595  & 0.789
& 86.2  & 0.357  & 1.004  & 0.551
& 86.8  & 0.411  & 0.615  & 0.476 \\
Uncertainty
& 57.3  & 0.000  & 66.30 & 3.987
& 70.1  & 0.000  & 19.46 & 4.726
& 71.4  & 0.000  & 17.86 & 4.808
& 77.9  & 0.010  & 5.007  & 1.777
& 82.0  & 0.075  & 1.750  & 1.069
& 86.4  & 0.190  & 0.379  & 0.665
& 87.4  & 0.226  & 0.237  & 0.584 \\
Margin
& 62.4  & 0.267  & 0.894  & 1.361
& 94.7  & 0.362  & 0.115  & 0.462
& 100 & 0.333  & 0.137  & 0.479
& 85.4 & 0.348  & 0.276  & 0.618
& 85.5  & 0.350  & 0.266  & 0.612
& 87.3  & 0.398  & 0.119  & 0.508
& 87.7  & 0.416  & 0.087  & 0.475 \\
Entropy
& 72.0  & 0.000  & 41.39 & 4.450
& 57.4  & 0.000  & 57.33 & 4.091
& 69.0  & 0.000  & 21.11 & 4.556
& 73.4  & 0.000  & 20.41 & 3.828
& 78.6  & 0.016  & 3.898  & 1.433
& 86.0  & 0.134  & 0.188  & 0.690
& 87.3  & 0.161  & 0.105  & 0.614 \\
TypiClust
& 72.7  & 0.201  & 4.324  & 1.445
& 100 & 0.647  & 0.529  & 0.140
& 82.3  & 0.734  & 1.142  & 0.291
& 82.8  & 0.752  & 0.892  & 0.256
& 84.3  & 0.760  & 0.643  & 0.212
& - & - & - & -
& - & - & - & - \\
DBAL
& 61.5  & 0.000  & 61.75 & 4.078
& 55.4  & 0.000  & 51.67 & 4.223
& 70.4  & 0.000  & 23.07 & 4.499
& 71.9  & 0.000  & 25.99 & 4.604
& 78.5  & 0.037  & 2.457  & 1.183
& 86.2  & 0.149  & 0.162  & 0.647
& 87.4  & 0.172  & 0.101  & 0.588 \\
\bottomrule
\end{tabular}
}
\end{table*}

\begin{table*}[ht]
\centering
\caption{PALM parameter estimates for CIFAR-10 using SimCLR embeddings for feature extraction, evaluated across various AL strategies and different numbers of labeled points used for curve fitting based on the maximum values from 5 repetitions. The table reports the maximum achievable accuracy ($\text{A}_{\max}$), coverage efficiency ($\delta$), early-stage performance offset ($\alpha$), and scalability ($\beta$). The results highlight the acceleration of learning dynamics with pretrained embeddings, where TypiClust and Margin benefit from high $\delta$ and low $\alpha$, indicating efficient early-stage learning. Conversely, methods like Entropy and DBAL exhibit delayed improvements at small budgets but show recovery and better performance as the annotation grows.}
\label{tab:cifar10_simclr_params3}
\resizebox{\textwidth}{!}{
\begin{tabular}{lcccccccccccccccccccccccccccc}
\toprule
\multirow{2}{*}{\textbf{AL Method}} 
& \multicolumn{4}{c}{\textbf{6 Points}} 
& \multicolumn{4}{c}{\textbf{10 Points}} 
& \multicolumn{4}{c}{\textbf{20 Points}} 
& \multicolumn{4}{c}{\textbf{50 Points}} 
& \multicolumn{4}{c}{\textbf{100 Points}} 
& \multicolumn{4}{c}{\textbf{500 Points}} 
& \multicolumn{4}{c}{\textbf{1000 Points}} \\
\cmidrule(lr){2-5} \cmidrule(lr){6-9} \cmidrule(lr){10-13} \cmidrule(lr){14-17} \cmidrule(lr){18-21} \cmidrule(lr){22-25} \cmidrule(lr){26-29}
& $\text{A}_{\max}$ & $\delta$ & $\alpha$ & $\beta$
& $\text{A}_{\max}$ & $\delta$ & $\alpha$ & $\beta$
& $\text{A}_{\max}$ & $\delta$ & $\alpha$ & $\beta$
& $\text{A}_{\max}$ & $\delta$ & $\alpha$ & $\beta$
& $\text{A}_{\max}$ & $\delta$ & $\alpha$ & $\beta$
& $\text{A}_{\max}$ & $\delta$ & $\alpha$ & $\beta$
& $\text{A}_{\max}$ & $\delta$ & $\alpha$ & $\beta$ \\
\midrule
Random
& 100 & 0.509  & 0.559  & 0.321
& 100 & 0.518  & 0.494  & 0.306
& 96.5  & 0.531  & 0.535  & 0.327
& 84.9  & 0.480  & 1.317  & 0.556
& 85.3 & 0.511  & 1.096  & 0.511
& 86.9  & 0.626  & 0.362  & 0.342
& 87.3  & 0.652  & 0.238  & 0.303 \\
Uncertainty
& 39.5  & 0.000  & 6.601  & 5.378
& 85.5  & 0.000  & 36.72 & 4.202
& 80.0  & 0.000  & 29.09 & 4.311
& 81.3  & 0.000  & 23.79 & 3.425
& 83.1  & 0.009  & 9.458  & 1.690
& 86.7  & 0.246  & 1.170  & 0.625
& 87.7  & 0.322  & 0.564  & 0.503\\
Margin
& 100 & 0.026  & 5.817  & 1.720
& 79.0  & 0.000  & 14.39 & 4.978
& 81.4  & 0.117  & 3.350  & 1.421
& 84.3  & 0.461  & 1.076  & 0.665
& 85.7  & 0.546  & 0.641  & 0.508
& 87.7  & 0.634  & 0.232  & 0.338
& 88.1  & 0.652  & 0.172  & 0.308 \\
Entropy
& 45.9  & 0.000  & 14.25 & 5.363
& 43.3  & 0.000  & 12.78 & 5.494
& 88.4  & 0.000  & 38.05 & 4.203
& 78.6  & 0.000  & 32.07 & 4.160
& 80.9  & 0.000  & 19.07 & 2.554
& 86.4  & 0.268  & 0.739  & 0.541
& 87.7  & 0.320  & 0.404  & 0.456 \\
TypiClust
& 78.3  & 0.000   & 10.77 & 6.109
& 78.3  & 0.000   & 10.73 & 6.118
& 100 & 0.741   & 0.017  & 0.078
& 93.5  & 0.793   & 0.023  & 0.092
& 88.2  & 0.836   & 0.048  & 0.124
& - & - & - & -
& - & - & - & - \\
DBAL
& 46.5  & 0.000  & 15.57 & 5.270
& 41.8  & 0.000  & 12.80 & 5.467
& 87.1  & 0.000  & 36.19 & 4.217
& 78.5  & 0.000  & 31.29 & 4.163
& 81.1  & 0.000  & 20.40 & 2.700
& 86.3  & 0.243  & 0.901  & 0.576
& 87.5  & 0.304  & 0.454  & 0.474 \\
\bottomrule
\end{tabular}
}
\end{table*}

\begin{table*}[ht]
\centering
\caption{PALM parameter estimates for CIFAR-100 across different AL methods and varying numbers of labeled points used for estimation. The table reports the maximum achievable accuracy ($\text{A}_{\max}$), coverage efficiency ($\delta$), early-stage performance offset ($\alpha$), and scalability ($\beta$). Higher values of $\delta$ and lower values of $\alpha$ indicate more efficient early-stage learning, while $\beta$ reflects the scalability of the method as the number of labeled points increases.}
\label{tab:cifar100_params}
\resizebox{\textwidth}{!}{
\begin{tabular}{lcccccccccccccccccccccccc}
\toprule
\multirow{2}{*}{\textbf{AL Method}} 
& \multicolumn{4}{c}{\textbf{6 Points}} 
& \multicolumn{4}{c}{\textbf{10 Points}} 
& \multicolumn{4}{c}{\textbf{20 Points}} 
& \multicolumn{4}{c}{\textbf{50 Points}} 
& \multicolumn{4}{c}{\textbf{100 Points}} 
& \multicolumn{4}{c}{\textbf{448 Points}} \\
\cmidrule(lr){2-5} \cmidrule(lr){6-9} \cmidrule(lr){10-13} \cmidrule(lr){14-17} \cmidrule(lr){18-21} \cmidrule(lr){22-25}
& $\text{A}_{\max}$ & $\delta$ & $\alpha$ & $\beta$
& $\text{A}_{\max}$ & $\delta$ & $\alpha$ & $\beta$
& $\text{A}_{\max}$ & $\delta$ & $\alpha$ & $\beta$
& $\text{A}_{\max}$ & $\delta$ & $\alpha$ & $\beta$
& $\text{A}_{\max}$ & $\delta$ & $\alpha$ & $\beta$
& $\text{A}_{\max}$ & $\delta$ & $\alpha$ & $\beta$ \\
\midrule
Random 
& 59.0 & 0.000 & 80.98 & 3.804
& 52.3 & 0.000 & 68.27 & 3.935
& 55.5 & 0.000 & 85.53 & 3.707
& 36.4 & 0.081 & 1.376 & 0.675
& 79.4 & 0.050 & 0.720 & 0.518
& 79.8 & 0.048 & 0.955 & 0.526 \\
Uncertainty 
& 55.6 & 0.000 & 86.03 & 3.771
& 43.7 & 0.000 & 85.64 & 3.781
& 46.1 & 0.000 & 85.11 & 3.715
& 93.8 & 0.035 & 0.410 & 0.531
& 100 & 0.030 & 0.626 & 0.551
& 69.9 & 0.029 & 3.469 & 0.656 \\
Margin 
& 44.8 & 0.000 & 85.36 & 3.772
& 56.6 & 0.000 & 85.26 & 3.748
& 46.6 & 0.000 & 86.15 & 3.728
& 25.8 & 0.000 & 73.40 & 3.936
& 100 & 0.042 & 1.044 & 0.482
& 58.0 & 0.025 & 10.64 & 0.751 \\
Entropy 
& 56.1 & 0.000 & 84.65 & 3.783
& 40.2 & 0.000 & 97.83 & 3.668
& 56.1 & 0.000 & 83.88 & 3.746
& 23.2 & 0.000 & 70.11 & 4.037
& 100 & 0.025 & 0.772 & 0.568
& 61.3 & 0.018 & 7.519 & 0.777 \\
TypiClust 
& 52.0 & 0.000 & 90.99 & 3.712
& 68.1 & 0.000 & 67.31 & 3.953
& 54.5 & 0.000 & 86.69 & 3.737
& - & - & - & -
& - & - & - & -
& - & - & - & - \\
DBAL 
& 51.3 & 0.000 & 88.04 & 3.764
& 52.6 & 0.000 & 88.59 & 3.740
& 54.7 & 0.000 & 84.73 & 3.736
& 23.3 & 0.000 & 54.62 & 4.185
& 47.5 & 0.031 & 1.284 & 0.774
& 100 & 0.024 & 0.452 & 0.595 \\
\bottomrule
\end{tabular}
}
\end{table*}

\begin{table*}[ht]
\centering
\caption{PALM parameter estimates for CIFAR-100 using SimCLR embeddings, evaluated across different AL strategies and varying numbers of labeled points used for estimation. The table reports the maximum achievable accuracy ($\text{A}_{\max}$), coverage efficiency ($\delta$), early-stage performance offset ($\alpha$), and scalability ($\beta$). Higher $\delta$ values and lower $\alpha$ indicate more efficient early-stage learning, while $\beta$ captures the rate of accuracy growth as the number of labeled points increases.}
\label{tab:cifar100_simclr_params}
\resizebox{\textwidth}{!}{
\begin{tabular}{lcccccccccccccccccccccccc}
\toprule
\multirow{2}{*}{\textbf{AL Method}} 
& \multicolumn{4}{c}{\textbf{6 Points}} 
& \multicolumn{4}{c}{\textbf{10 Points}} 
& \multicolumn{4}{c}{\textbf{20 Points}} 
& \multicolumn{4}{c}{\textbf{50 Points}} 
& \multicolumn{4}{c}{\textbf{100 Points}} 
& \multicolumn{4}{c}{\textbf{448 Points}} \\
\cmidrule(lr){2-5} \cmidrule(lr){6-9} \cmidrule(lr){10-13} \cmidrule(lr){14-17} \cmidrule(lr){18-21} \cmidrule(lr){22-25}
& $\text{A}_{\max}$ & $\delta$ & $\alpha$ & $\beta$
& $\text{A}_{\max}$ & $\delta$ & $\alpha$ & $\beta$
& $\text{A}_{\max}$ & $\delta$ & $\alpha$ & $\beta$
& $\text{A}_{\max}$ & $\delta$ & $\alpha$ & $\beta$
& $\text{A}_{\max}$ & $\delta$ & $\alpha$ & $\beta$
& $\text{A}_{\max}$ & $\delta$ & $\alpha$ & $\beta$ \\
\midrule
Random 
& 41.7 & 0.201 & 0.718 & 1.133
& 47.7 & 0.205 & 0.570 & 0.975
& 49.9 & 0.218 & 0.458 & 0.887
& 53.9 & 0.256 & 0.227 & 0.709
& 55.7 & 0.278 & 0.138 & 0.626
& 57.6 & 0.318 & 0.057 & 0.516 \\
Uncertainty 
& 51.0 & 0.000 & 88.128 & 3.777
& 57.3 & 0.000 & 83.024 & 3.805
& 38.4 & 0.037 & 0.615 & 1.347
& 46.6 & 0.059 & 0.000 & 1.033
& 51.3 & 0.073 & 0.000 & 0.881
& 58.2 & 0.116 & 0.001 & 0.634 \\
Margin 
& 56.0 & 0.000 & 65.254 & 4.041
& 39.9 & 0.032 & 2.265 & 1.822
& 46.9 & 0.104 & 0.969 & 1.157
& 52.6 & 0.163 & 0.366 & 0.840
& 55.3 & 0.195 & 0.175 & 0.707
& 58.0 & 0.237 & 0.068 & 0.573 \\
Entropy 
& 59.6 & 0.000 & 73.883 & 3.920
& 60.1 & 0.000 & 77.654 & 3.865
& 48.1 & 0.000 & 87.174 & 3.733
& 42.4 & 0.030 & 0.000 & 1.227
& 47.3 & 0.047 & 0.000 & 0.991
& 58.6 & 0.089 & 0.001 & 0.651 \\
TypiClust 
& 45.9 & 0.433 & 1.128 & 0.737
& 48.3 & 0.451 & 0.948 & 0.633
& 48.8 & 0.464 & 0.860 & 0.597
& - & - & - & -
& - & - & - & -
& - & - & - & - \\
DBAL 
& 59.6 & 0.000 & 73.29 & 3.937
& 60.0 & 0.000 & 77.83 & 3.864
& 47.4 & 0.000 & 87.58 & 3.726
& 42.0 & 0.030 & 0.000 & 1.246
& 47.3 & 0.048 & 0.000 & 0.989
& 58.6 & 0.091 & 0.001 & 0.646 \\
\bottomrule
\end{tabular}
}
\end{table*}

\clearpage

\begin{table*}[ht]
\centering
\caption{PALM parameter estimates on ImageNet-50 across different AL strategies and self-supervised embeddings. The table reports the maximum achievable accuracy ($\text{A}_{\max}$), coverage efficiency ($\delta$), early-stage performance offset ($\alpha$), and scalability ($\beta$) for each method, capturing the efficiency and dynamics of learning across varying annotation budgets.}
\label{tab:imagenet50_full_params}
\resizebox{\textwidth}{!}{
\begin{tabular}{lcccccccccccccccccccccccc}
\toprule
\multirow{2}{*}{\textbf{Embedding}} 
& \multicolumn{4}{c}{\textbf{Random}} 
& \multicolumn{4}{c}{\textbf{Entropy}} 
& \multicolumn{4}{c}{\textbf{Uncertainty}} 
& \multicolumn{4}{c}{\textbf{DBAL}} 
& \multicolumn{4}{c}{\textbf{Margin}} 
& \multicolumn{4}{c}{\textbf{TypiClust}} \\
\cmidrule(lr){2-5} \cmidrule(lr){6-9} \cmidrule(lr){10-13} \cmidrule(lr){14-17} \cmidrule(lr){18-21} \cmidrule(lr){22-25}
 & $\text{A}_{\max}$ & $\delta$ & $\alpha$ & $\beta$
 & $\text{A}_{\max}$ & $\delta$ & $\alpha$ & $\beta$
 & $\text{A}_{\max}$ & $\delta$ & $\alpha$ & $\beta$
 & $\text{A}_{\max}$ & $\delta$ & $\alpha$ & $\beta$
 & $\text{A}_{\max}$ & $\delta$ & $\alpha$ & $\beta$
 & $\text{A}_{\max}$ & $\delta$ & $\alpha$ & $\beta$ \\
\midrule
MoCov2+ 
& 27.5 & 0.005 & 4.64 & 1.86
& 47.6 & 0.006 & 2.27 & 2.36
& 1.95 & 1.000 & 0.00 & -1.00
& 66.3 & 0.080 & 13.25 & -0.17
& 27.1 & 0.000 & 17.77 & 4.49
& 28.4 & 0.000 & 17.63 & 4.91 \\
MoCov3 
& 87.0 & 0.397 & 0.00 & 0.78
& 84.5 & 0.155 & 0.00 & 1.20
& 85.4 & 0.154 & 0.41 & 1.22
& 84.6 & 0.103 & 0.61 & 1.39
& 88.2 & 0.444 & 0.00 & 0.78
& 87.0 & 0.714 & 0.00 & 0.55 \\
BYOL 
& 8.48 & 0.314 & 0.00 & 0.67
& 42.0 & 0.071 & 0.00 & 0.30
& 12.0 & 0.170 & 0.00 & 0.61
& 23.1 & 0.129 & 0.00 & 0.31
& 8.08 & 0.250 & 0.43 & 0.81
& 2.96 & 0.130 & 6.56 & 1.05 \\
SimCLR 
& 66.5 & 0.131 & 0.22 & 0.98
& 45.9 & 0.001 & 3.81 & 2.39
& 50.0 & 0.009 & 1.11 & 1.96
& 46.9 & 0.001 & 4.42 & 2.71
& 62.7 & 0.138 & 0.00 & 1.15
& 59.5 & 0.363 & 1.13 & 0.67 \\
\bottomrule
\end{tabular}
}
\end{table*}

\begin{table*}[ht]
\centering
\caption{PALM parameter estimates on ImageNet-100 across different AL strategies and self-supervised embeddings. The table reports the maximum achievable accuracy ($\text{A}_{\max}$), coverage efficiency ($\delta$), early-stage performance offset ($\alpha$), and scalability ($\beta$) for each method, providing insights into learning efficiency and model behavior across varying annotation budgets.}
\label{tab:imagenet100_full_params}
\resizebox{\textwidth}{!}{
\begin{tabular}{lcccccccccccccccccccccccc}
\toprule
\multirow{2}{*}{\textbf{Embedding}} 
& \multicolumn{4}{c}{\textbf{Random}} 
& \multicolumn{4}{c}{\textbf{Entropy}} 
& \multicolumn{4}{c}{\textbf{Uncertainty}} 
& \multicolumn{4}{c}{\textbf{DBAL}} 
& \multicolumn{4}{c}{\textbf{Margin}} 
& \multicolumn{4}{c}{\textbf{TypiClust}} \\
\cmidrule(lr){2-5} \cmidrule(lr){6-9} \cmidrule(lr){10-13} \cmidrule(lr){14-17} \cmidrule(lr){18-21} \cmidrule(lr){22-25}
 & $\text{A}_{\max}$ & $\delta$ & $\alpha$ & $\beta$
 & $\text{A}_{\max}$ & $\delta$ & $\alpha$ & $\beta$
 & $\text{A}_{\max}$ & $\delta$ & $\alpha$ & $\beta$
 & $\text{A}_{\max}$ & $\delta$ & $\alpha$ & $\beta$
 & $\text{A}_{\max}$ & $\delta$ & $\alpha$ & $\beta$
 & $\text{A}_{\max}$ & $\delta$ & $\alpha$ & $\beta$ \\
\midrule
MoCov2+ 
& 21.2 & 0.001 & 6.51 & 2.26
& 1.85 & 0.306 & 0.00 & 6.75
& 100 & 0.002 & 20.0 & 0.39
& 2.49 & 0.328 & 0.00 & 2.07
& 21.4 & 0.055 & 0.00 & 1.26
& 27.60 & 0.001 & 20.0 & 2.26 \\
MoCov3 
& 81.6 & 0.333 & 0.00 & 0.87
& 76.7 & 0.082 & 0.00 & 1.47
& 76.6 & 0.112 & 0.00 & 1.37
& 76.4 & 0.083 & 0.00 & 1.48
& 81.1 & 0.316 & 0.00 & 1.00
& 80.7 & 0.690 & 0.00 & 0.56 \\
BYOL 
& 8.17 & 0.188 & 0.00 & 0.47
& 8.52 & 0.125 & 0.34 & 0.60
& 22.7 & 0.070 & 0.00 & 0.36
& 8.64 & 0.090 & 1.71 & 0.72
& 4.44 & 0.302 & 0.00 & 0.72
& 2.46 & 0.000 & 20.0 & 4.41 \\
SimCLR 
& 44.0 & 0.226 & 0.00 & 0.73
& 27.7 & 0.087 & 0.00 & 1.08
& 36.2 & 0.105 & 0.00 & 0.86
& 27.3 & 0.089 & 0.00 & 1.09
& 44.0 & 0.171 & 0.30 & 0.80
& 36.9 & 0.090 & 3.36 & 1.20 \\
\bottomrule
\end{tabular}
}
\end{table*}

\begin{table*}[ht]
\centering
\caption{PALM parameter estimates on ImageNet-200 across different AL strategies and self-supervised embeddings. The table reports the maximum achievable accuracy ($\text{A}_{\max}$), coverage efficiency ($\delta$), early-stage performance offset ($\alpha$), and scalability ($\beta$) for each method, highlighting differences in learning dynamics and sample efficiency across strategies.}
\label{tab:imagenet200_full_params}
\resizebox{\textwidth}{!}{
\begin{tabular}{lcccccccccccccccccccccccc}
\toprule
\multirow{2}{*}{\textbf{Embedding}} 
& \multicolumn{4}{c}{\textbf{Random}} 
& \multicolumn{4}{c}{\textbf{Entropy}} 
& \multicolumn{4}{c}{\textbf{Uncertainty}} 
& \multicolumn{4}{c}{\textbf{DBAL}} 
& \multicolumn{4}{c}{\textbf{Margin}} 
& \multicolumn{4}{c}{\textbf{TypiClust}} \\
\cmidrule(lr){2-5} \cmidrule(lr){6-9} \cmidrule(lr){10-13} \cmidrule(lr){14-17} \cmidrule(lr){18-21} \cmidrule(lr){22-25}
 & $\text{A}_{\max}$ & $\delta$ & $\alpha$ & $\beta$
 & $\text{A}_{\max}$ & $\delta$ & $\alpha$ & $\beta$
 & $\text{A}_{\max}$ & $\delta$ & $\alpha$ & $\beta$
 & $\text{A}_{\max}$ & $\delta$ & $\alpha$ & $\beta$
 & $\text{A}_{\max}$ & $\delta$ & $\alpha$ & $\beta$
 & $\text{A}_{\max}$ & $\delta$ & $\alpha$ & $\beta$ \\
\midrule
MoCov2+ 
& 18.9 & 0.001 & 4.65 & 2.33
& 1.12 & 0.411 & 0.00 & 1.05
& 3.90 & 0.000 & 18.9 & 4.63
& 1.40 & 0.000 & 5.44 & 5.17
& 32.0 & 0.020 & 1.41 & 1.62
& 13.1 & 0.060 & 3.57 & 1.85 \\
MoCov3 
& 76.9 & 0.330 & 0.00 & 0.80
& 47.4 & 0.092 & 0.00 & 1.38
& 70.3 & 0.129 & 0.00 & 1.28
& 69.2 & 0.087 & 0.00 & 1.43
& 76.0 & 0.348 & 0.00 & 0.83
& 70.4 & 0.694 & 0.00 & 0.57 \\
BYOL 
& 27.0 & 0.034 & 0.01 & 0.40
& 14.1 & 0.068 & 0.00 & 0.39
& 3.29 & 0.206 & 0.00 & 0.71
& 7.20 & 0.136 & 0.00 & 0.42
& 7.98 & 0.095 & 0.00 & 0.50
& 28.5 & 0.000 & 20.8 & 4.08 \\
SimCLR 
& 36.6 & 0.027 & 1.59 & 1.48
& 19.8 & 0.005 & 1.56 & 1.85
& 30.1 & 0.011 & 0.70 & 1.54
& 18.3 & 0.002 & 2.76 & 2.18
& 37.9 & 0.057 & 0.36 & 1.26
& 42.0 & 0.378 & 0.00 & 0.41 \\
\bottomrule
\end{tabular}
}
\end{table*}

\end{document}